\documentclass{article}

\PassOptionsToPackage{dvipsnames}{xcolor}

\usepackage[accepted]{icml2025}
\usepackage{hyperref}
\usepackage{microtype}   
\usepackage{graphicx}    
\usepackage{booktabs}    
\usepackage{amsmath}     
\usepackage{amssymb}       
\usepackage{amsthm}
\usepackage{adjustbox}
\usepackage{enumitem} 
\usepackage{float}
\usepackage{multirow}   
\usepackage{makecell} 
\usepackage{subcaption}

\definecolor{BlueViolet}{HTML}{8A2BE2}
\hypersetup{colorlinks=true,
            linkcolor=RoyalPurple, 
            citecolor=RoyalPurple, 
            urlcolor=RoyalPurple,
            pdftitle={Graph-Assisted Stitching for Offline Hierarchical Reinforcement Learning},
            pdfauthor  = {Seungho Baek, Taegeon Park, Jongchan Park, Seungjun Oh, Yusung Kim},
            pdfsubject = {Offline Hierarchical Reinforcement Learning},
            pdfkeywords= {Offline Hierarchical Reinforcement Learning, Offline Goal-Conditioned Reinforcement Learning, Graph-based Reinforcement Learning, Temporal Distance Representation Learning}}

\icmltitlerunning{Graph-Assisted Stitching for Offline Hierarchical Reinforcement Learning}

\begin{document}
\frenchspacing 
\twocolumn[
\icmltitle{Graph-Assisted Stitching for Offline Hierarchical Reinforcement Learning}
\begin{icmlauthorlist}
\icmlauthor{Seungho Baek}{cse}
\icmlauthor{Taegeon Park}{ai}
\icmlauthor{Jongchan Park}{ai}
\icmlauthor{Seungjun Oh}{ai}
\icmlauthor{Yusung Kim}{cse,ai}
\end{icmlauthorlist}
\icmlaffiliation{cse}{Department of Computer Science and Engineering, Sungkyunkwan University, Suwon, Republic of Korea} 
\icmlaffiliation{ai}{Department of Artificial Intelligence, Sungkyunkwan University, Suwon, Republic of Korea}
\icmlcorrespondingauthor{Yusung Kim}{yskim525@skku.edu}
\vskip 0.3in
]
\printAffiliationsAndNotice{}

\begin{abstract}
Existing offline hierarchical reinforcement learning methods rely on high-level policy learning to generate subgoal sequences. However, their efficiency degrades as task horizons increase, and they lack effective strategies for stitching useful state transitions across different trajectories. We propose Graph-Assisted Stitching (GAS), a novel framework that formulates subgoal selection as a graph search problem rather than learning an explicit high-level policy. By embedding states into a Temporal Distance Representation (TDR) space, GAS clusters semantically similar states from different trajectories into unified graph nodes, enabling efficient transition stitching. A shortest-path algorithm is then applied to select subgoal sequences within the graph, while a low-level policy learns to reach the subgoals. To improve graph quality, we introduce the Temporal Efficiency (TE) metric, which filters out noisy or inefficient transition states, significantly enhancing task performance. GAS outperforms prior offline HRL methods across locomotion, navigation, and manipulation tasks. Notably, in the most stitching-critical task, it achieves a score of 88.3, dramatically surpassing the previous state-of-the-art score of 1.0. Our source code is available at: \url{https://github.com/qortmdgh4141/GAS}.
\end{abstract}

\section{Introduction}
Offline reinforcement learning (RL) has emerged as a promising approach to learning effective policies from pre-collected datasets without requiring active interaction with the environment~\cite{lange2012batch, kumar2019stabilizing}. This paradigm is particularly valuable in real-world applications where online data collection is costly, time-consuming, or even unsafe (e.g., robotics, healthcare, and autonomous driving)~\cite{gulcehre2020rlunplugged, jeong2024prediction}. By leveraging offline datasets, agents can learn optimal behaviors while avoiding risks associated with real-world deployment. Building on this paradigm, offline goal-conditioned reinforcement learning (GCRL) provides a general framework for learning a multi-task policy from diverse data. Nevertheless, long-horizon and sparse-reward tasks remain a fundamental challenge in offline GCRL~\cite{ajay2021opal, li2024boosting}. In such tasks, rewards are often provided only at the final state, making it difficult to establish a clear learning signal for intermediate decision-making. Without explicit guidance on how to decompose long-horizon goals into manageable subgoals, standard offline GCRL algorithms struggle to generalize and effectively utilize the available data. This limitation has motivated research into offline hierarchical reinforcement learning (HRL). HRL is a well-established approach for tackling long-horizon and sparse-reward problems by introducing a two-level decision-making framework~\cite{vezhnevets2017feudal, nachum2018data, levy2019learning}: A high-level policy generates subgoals that guide long-term planning. A low-level policy learns how to achieve these subgoals using primitive actions. By breaking down complex tasks into subgoal-conditioned learning problems, HRL can significantly improve learning efficiency. 

\begin{figure*}[t]
    \centering
    \includegraphics[width=1.0\textwidth]{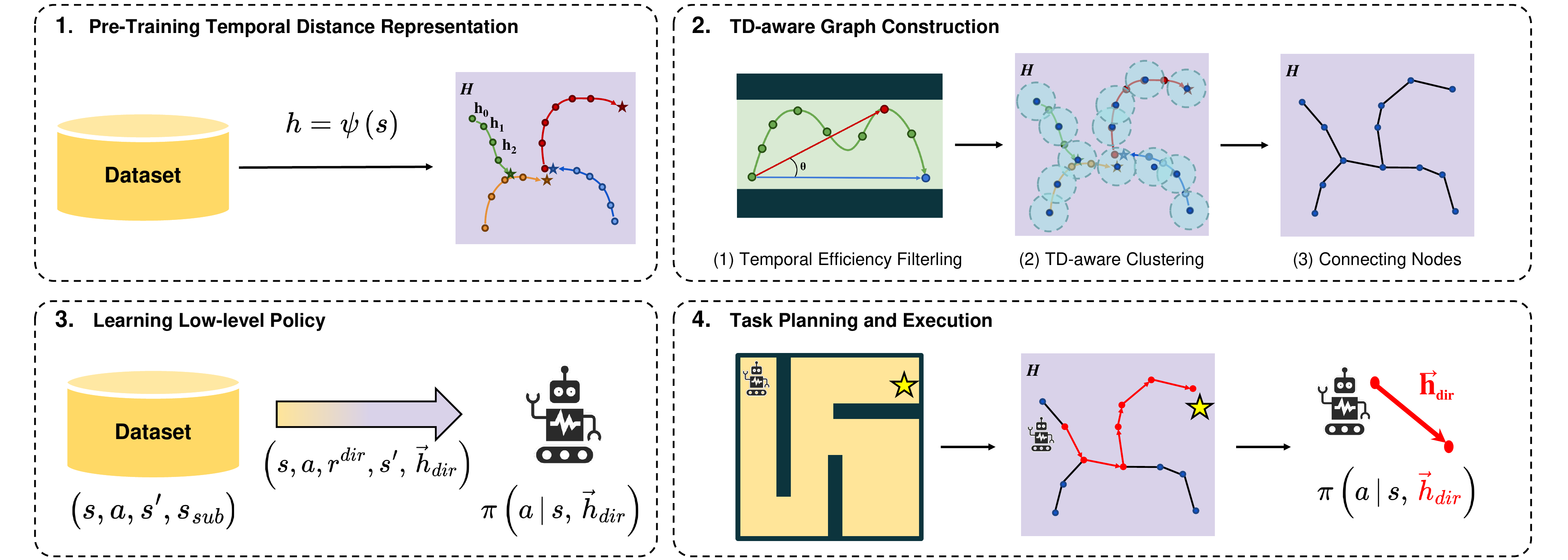} 
    \caption{\textbf{Overview of GAS.} (1) A Temporal Distance Representation (TDR) is pretrained from an offline dataset. (2) A graph is constructed by selecting only high-TE states based on the Temporal Efficiency (TE) metric. (3) A TD-based subgoal-conditioned low-level policy is trained using all states. (4) The graph is utilized for task planning and subgoal selection, while action execution is performed by the low-level policy.}
    \label{fig:overview}
\end{figure*}

Despite its potential, existing offline HRL methods exhibit major limitations primarily in high-level policy learning, leading to critical challenges: \textbf{(1) Lack of Temporal Awareness in Subgoal Sampling:} Most approaches sample subgoal candidates by choosing states at fixed time intervals along a given trajectory~\cite{park2023hiql, shin2023guide}. However, this naive subgoal sampling strategy fails to capture the temporal dependencies between states, leading to redundant or inefficient subgoals that hinder high-level policy learning. In offline datasets, where suboptimal behaviors and noisy actions are common, such heuristics may misinterpret temporal transitions, resulting in unstable high-level planning. \textbf{(2) Challenges in Long-Horizon Reasoning:} As reported in recent studies, prior works achieve strong performance in medium-scale environments such as antmaze-medium but fail to scale to larger environments like antmaze-giant, where the task horizon is significantly longer~\cite{park2025ogbench, sobal2025latent}. As task horizons grow longer, sparse rewards cause the learning signal to weaken, increasing noise and reducing the effectiveness of subgoal generation by high-level policy learning. \textbf{(3) Ineffective Trajectory Stitching:} Offline datasets often consist of diverse trajectories collected attempting to achieve various goals~\cite{park2025ogbench, park2025flow}. Effective offline HRL should be able to stitch together meaningful state transitions across different trajectories. However, existing methods lack efforts in cross-goal stitching, where partial segments from different goal-oriented trajectories are reused to form new trajectory compositions that facilitate goal generalization. 

To address these limitations, we propose Graph-Assisted Stitching (GAS), a novel framework that adopts a graph-based subgoal selection approach without the need for explicit high-level policy learning for subgoal generation. GAS leverages a pretrained Temporal Distance Representation (TDR) to estimate the minimum number of time steps required for an optimal policy to transition between states. States in the dataset are clustered at regular temporal distance intervals, and the centroids of each cluster are treated as graph nodes. Edges are created between nodes if their temporal distance falls below a predefined threshold. By constructing this graph, each node groups semantically similar states, while edges enable stitching between states that were not directly connected in the dataset, significantly improving  the effectiveness of transition stitching. Additionally, we introduce the Temporal Efficiency (TE) metric to enhance the quality of the graph. TE evaluates how close a state’s transition is to the optimal temporal behavior. Only states with high TE values are selected for graph construction, reducing computational overhead while preserving high-quality subgoal selection. 

The key contributions of this work are as follows:
\begin{itemize}[leftmargin=1.0em, itemsep=0.5em, topsep=3.0pt]
\item We propose a novel framework for offline HRL that leverages a graph-based approach for subgoal selection to enable efficient long-horizon reasoning and state transition stitching.
\item We introduce TD-aware graph construction that groups semantically similar states while stitching meaningful state transitions across trajectories.
\item We introduce the TE metric, which filters out inefficient transition states to enhance graph quality and reduce construction overhead.
\item We evaluate GAS across diverse numeric and visual state environments, consistently outperforming baselines and achieving remarkable improvements in long-horizon reasoning and trajectory stitching tasks.
\end{itemize}

\section{Related Work}
\subsection{Offline Goal-Conditioned RL}
Offline goal-conditioned reinforcement learning (GCRL) aims to learn a multi-task policy from diverse data without further environment interaction~\cite{ma2022gofar, yang2023essential, cao2024offline, sikchi2024smore, wang2024goplan}. Previous methods have proposed various approaches for training goal-conditioned value functions. These include hindsight relabeling, which relabels achieved states as goals for goal-conditioned learning~\cite{andrychowicz2017hindsight}; expectile regression for optimal value estimation~\cite{kostrikov2022offline, park2023hiql}; quasi-metric learning with a dual objective~\cite{wang2023optimal}; and contrastive representation learning to estimate returns and perform one-step policy improvement~\cite{eysenbach2022contrastive, zheng2024stabilizing}. However, these approaches often struggle with long-horizon and sparse-reward tasks, as rewards are typically provided only at the final state, making it difficult to establish a clear learning signal for intermediate decision-making. 

\subsection{Offline Hierarchical RL}
Offline hierarchical reinforcement learning (HRL) introduces a two-level decision-making framework, where a high-level policy generates subgoals and a low-level policy executes primitive actions to reach them. This structure enables temporal abstraction and decomposes complex tasks into more tractable sub-problems~\cite{haarnoja2018latent, nachum2019near, zhang2020generating}. Prior methods have proposed various techniques, including skill discovery methods to extract primitive skills for downstream learning~\cite{ajay2021opal, li2024boosting}, and a latent variable method for modeling the distribution of reachable subgoals~\cite{shin2023guide}. To extend offline HRL to high-dimensional or visual state environments, recent methods adopt subgoal representation learning in a latent space. HIQL~\cite{park2023hiql} propose a method where the high-level policy outputs latent subgoal representations learned from the value function. Similarly, HHILP~\cite{park2024foundation} produce subgoals in a Temporal Distance Representation (TDR) space, enhancing the generalization ability of the policy. Despite these advances, existing methods often struggle with high-level policy learning due to a lack of temporal awareness in subgoal sampling, weak learning signals in long-horizon tasks, and limited efforts in cross-goal stitching.

\subsection{Trajectory Stitching in Offline RL}
Stitching refers to the process of combining partial segments from different trajectories to form new trajectories that were not explicitly present in the original dataset~\cite{fu2020d4rl, park2025ogbench}. Previous offline reinforcement learning (RL) algorithms have primarily relied on implicit stitching through Bellman backup mechanisms~\cite{fujimoto2019off, wu2019behavior, kumar2020conservative, levine2020offline}. More recent approaches aim to improve an agent’s stitching ability by applying temporal goal augmentation techniques to supervised learning-based RL~\cite{ghugare2024temporal}, or by leveraging diffusion models to combine high-value trajectory segments from multiple trajectories~\cite{li2024diffstitch, kim2024ssd}. However, these approaches are typically tailored to single-task agents or low-dimensional goal spaces. As a result, they face structural limitations when extended to high-dimensional or visual state environments in multi-task settings. In contrast, our method performs explicit graph-based stitching and trains a multi-task agent by leveraging Temporal Distance Representations (TDR), which can scale to high-dimensional state environments.
 
\subsection{Graph Construction in Online RL}
Graph-based reinforcement learning (RL) constructs explicit graphs, where nodes represent goal states and edges are formed based on reachability. Prior works leverage such graphs to address the hierarchy tradeoff~\cite{lee2022dhrl} or to replace high-level policy learning with structured planning~\cite{eysenbach2019search, hoang2021successor}. For graph construction, existing approaches commonly sample candidate goals from the replay buffer and apply Farthest Point Sampling (FPS)~\cite{Eldar1994} to select representative nodes~\cite{kim2021landmark, lee2022dhrl, kim2023imitating, park2024ngte}, while others apply a coarse-to-fine grid refinement strategy over manually defined goal spaces~\cite{yoon2024breadth}. However, these methods are primarily designed for online RL to efficiently expand the graph, often measuring reachability through repeated online rollouts~\cite{faust2018prmrl, bagaria2020skill, yoon2024breadth}, which are not available in offline settings. FPS, a widely used technique for node selection, is highly sensitive to data distribution and often produces suboptimal graph structures. In addition, many previous methods rely on manually specified low-dimensional goal spaces, whereas our method constructs a graph in a latent space, enabling application to high-dimensional and visual state environments.

\section{Preliminaries}
\subsection{Goal-Conditioned MDP}
We consider continuous control tasks framed as a goal-conditioned Markov Decision Process (GC-MDP)~\cite{kaelbling1993learning}. Formally, each task is specified by the tuple \((\mathcal{S}, \mathcal{G}, \mathcal{A}, p, r, \gamma)\), where \(\mathcal{S}\) represents the state space, and we assume the goal space \(\mathcal{G}\) is identical to the state space (i.e., \(\mathcal{G} = \mathcal{S}\)). The set \(\mathcal{A}\) describes the action space, while \(p\) denotes the transition probability function \(p(s' \mid s, a)\).

Let \(r_t\) denote the reward at time step \(t\), and \(\gamma\) be the discount factor. The objective is to learn a policy \(\pi\) that maximizes the expected return:
\begin{equation}
J(\pi) = 
\mathbb{E}_{\pi} \left[ \sum_{t=0}^T \gamma^t \, r_t \right]
\end{equation}
In goal-conditioned environments, especially those with sparse rewards, the reward function often takes the following form:
\begin{equation}
r(s_t, a_t, s_{t+1}, g) =
\begin{cases}
1 & \text{if } \left\| \Omega(s_{t+1}) - \Omega(g) \right\|_2 < \epsilon \\
0 & \text{otherwise}
\end{cases}
\end{equation}
where \(\Omega(\cdot)\) denotes a projection function that extracts the goal-relevant components from the state or goal state. A reward of 1 is provided when the Euclidean distance between \(\Omega(s_{t+1)}\) and \(\Omega(g)\) is below the threshold \(\epsilon\), indicating proximity to the goal. Otherwise, the reward is set to 0.

\subsection{Hierarchical GC-MDPs}
This section extends the discussion from goal-conditioned Markov Decision Process (GC-MDP) to a hierarchical setting~\cite{kaelbling1993learning, sutton1999between, barto2003recent}. Specifically, existing hierarchical approaches decompose the decision-making process into two levels: a high-level policy $\pi^h$, which is responsible for generating subgoals that guide the agent toward the final goal, and a low-level policy $\pi^l$, which aims to reach these subgoals rather than the final goal directly. To formally describe the hierarchical structure, one can consider two coupled MDPs:

\textbf{High-Level MDP} 
$\mathcal{M}^h = (\mathcal{S}, \mathcal{G}_\text{final}, \mathcal{A}^h, p^h, r^h, \gamma)$: The action space $\mathcal{A}^h$ corresponds to selecting subgoals $g_\text{sub} \in \mathcal{G}_{\text{sub}}$. The high-level reward function $r^h$ provides feedback on how effectively the chosen subgoal promotes progress toward the final goal $g_\text{final} \in \mathcal{G}_\text{final}$.

\textbf{Low-Level MDP} 
$\mathcal{M}^l = (\mathcal{S}, \mathcal{G}_{\text{sub}}, \mathcal{A}^l, p^l, r^l, \gamma)$: The action space \(\mathcal{A}^l\) is identical to the original action space \(\mathcal{A}\), which consists of primitive actions \(a \in \mathcal{A}\). The low-level policy $\pi^l(a \mid s, g_\text{sub})$ is trained to maximize $r^l$, which measures how effectively the subgoal $g_\text{sub}$ is reached.

Sampling transitions for training hierarchical policies is crucial. Typically, a transition for the high-level policy can be represented as $(s_t, s_{t+c}, s_{g})$, where $s_t$ is the current state, $s_{t+c}$ is the state observed after $c$ steps, and $s_{g}$ is the final goal $g_\text{final}$ sampled from a future state within the same trajectory. The high-level policy is trained to generate $s_{t+c}$ as the subgoal $g_\text{sub}$. For the low-level policy, a common transition tuple is \((s_t, a, s_{t+1}, s_{t+c})\), where $a$ is the action taken by the policy, $s_{t+1}$ is the next state, and \(s_{t+c}\) is the subgoal $g_\text{sub}$ the low-level policy must achieve. In many existing methods~\cite{ajay2021opal, park2023hiql, shin2023guide, li2024boosting, park2024foundation}, the parameter $c$, which defines the subgoal horizon, is treated as a hyperparameter.

\subsection{Temporal Distance Representation Learning}
\begin{figure}[H]
    \centering
    \includegraphics[width=0.95\linewidth]{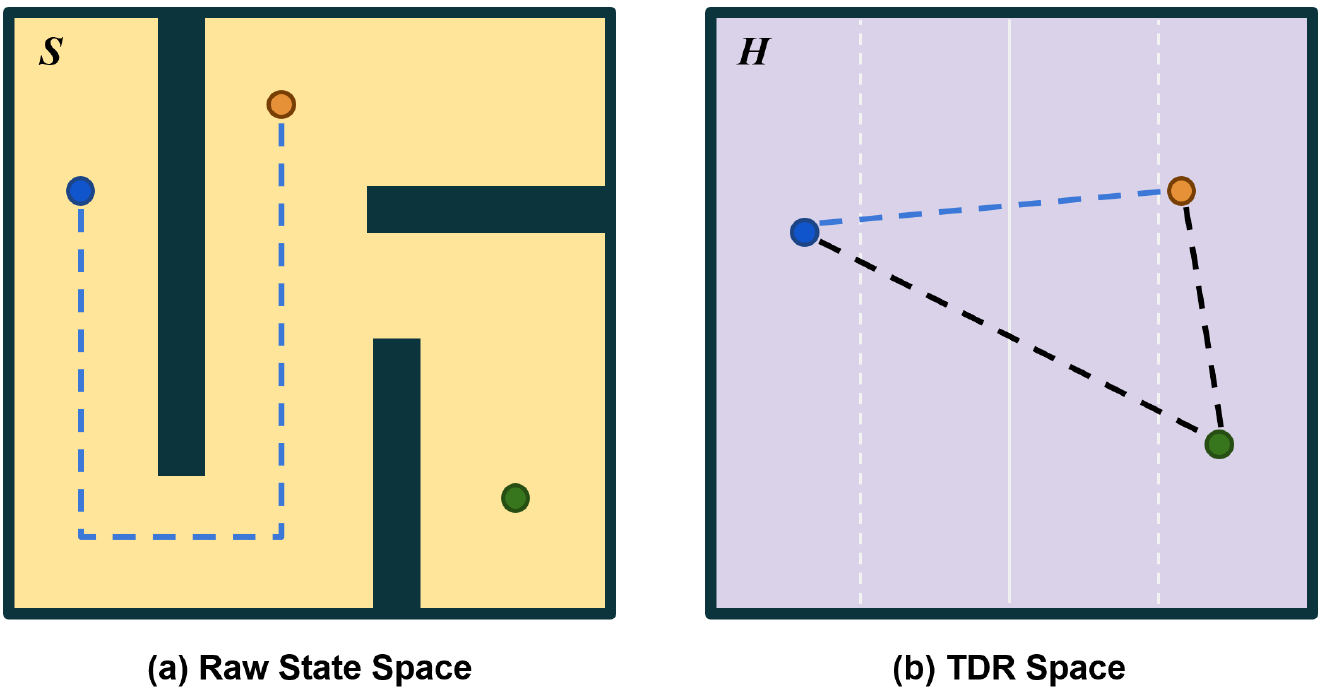}
    \caption{\textbf{Temporal Distance Representation (TDR).}
    \textit{(left)} Raw state space, where spatial proximity does not imply temporal closeness. \textit{(right)} TDR space, where Euclidean distances correspond to the minimum number of steps required to reach one state from another. This representation captures and preserves global temporal structure.}
    \label{fig:tdr}
\end{figure}

Recent work proposed a method to learn a Temporal Distance Representation (TDR) \(\psi\) from offline datasets for reinforcement learning (RL)~\cite{park2024foundation}. The central insight is to embed states into a latent space \(\mathcal{H}\). In this space, the Euclidean distance between any two points corresponds to the minimum number of steps required to transition from one state to another in the original state space \(\mathcal{S}\). (Figure~\ref{fig:tdr})

Formally, we aim for:
\begin{equation}
d^*(s, g) =
\bigl\|\psi(s) - \psi(g)\bigr\|_2
\end{equation}
where \(d^*(s, g)\) denotes the optimal temporal distance from \(s\) to \(g\). To learn \(\psi: \mathcal{S} \to \mathcal{H}\), one may view this as a goal-conditioned value function problem~\cite{kaelbling1993learning} and employ an existing offline RL algorithm such as IQL~\cite{kostrikov2022offline, park2023hiql}. 

Specifically, define:
\begin{equation}
V(s, g) =
-\,\bigl\|\psi(s) - \psi(g)\bigr\|_2
\end{equation}
By interpreting \(-\|\psi(s) - \psi(g)\|_2\) as a goal-conditioned value, we can optimize \(V\) via a temporal difference objective on offline data \(\mathcal{D}\) with the expectile loss \(\ell_\tau^2\)~\cite{newey1987asymmetric}. 
Accordingly, we define the TDR loss \(\mathcal{L}\) as:
\begin{equation}
\mathbb{E}_{(s,\,s',\,g)\,\sim\,\mathcal{D}}
\Bigl[\ell_{\tau}^2\Bigl(-{1}\{s \neq g\} + \gamma\,\overline{V}(s', g)  - V(s, g)\Bigr)\Bigr]
\end{equation}
\normalsize
\begin{equation}
\label{eq:expectile}
\ell_\tau^2 (x) = 
|\tau - 1(x < 0)| x^2
\end{equation}
where \(\overline{V}\) is a target value function, \(\gamma\) is the discount factor,  $\tau$ is the expectile coefficient, and \(-{1}\{s \neq g\}\) indicates that \(s\) differs from the goal \(g\). Through this procedure, the latent space \(\mathcal{H}\) preserves the optimal temporal distance in \(\mathcal{S}\), thus maintaining global long-horizon relationships among states.

\section{Graph-Assisted Stitching (GAS)}
This study proposes the GAS framework to overcome the limitations of offline Hierarchical Reinforcement Learning (HRL). GAS eliminates the need for high-level policy learning and instead adopts a graph-based approach to enable efficient long-horizon reasoning and state transition stitching. First, it constructs a graph using a clustering method that maintains consistent intervals in the Temporal Distance Representation (TDR) space, enabling stitching opportunities across diverse trajectories through connections between nodes. Then, it utilizes a shortest-path algorithm to sequentially select subgoals, while the low-level policy learns primitive actions to reach these subgoals. Finally, to address issues with low-quality graphs caused by suboptimal datasets, a Temporal Efficiency (TE) metric is introduced to not only enhance graph quality but also reduce construction overhead by filtering out ineffective states. (Figure~\ref{fig:overview})

\subsection{TD-Aware Graph Construction}
\label{sec:td_aware_graph_construction}
To construct the graph, all states in the dataset are clustered in the Temporal Distance Representation (TDR) space at intervals of the target temporal distance $H_\text{TD}$. Initially, the first state of the dataset is assigned as the center of the first cluster. For each subsequent state, we compute its temporal distance to the existing cluster centers and assign it to the nearest cluster if the distance is within $H_\text{TD}$. Otherwise, a new cluster is created with the state as its center. Once all states are clustered, the cluster centers are updated to the mean values of the states within each cluster. This approach effectively covers all states in the dataset while maintaining consistent spacing between clusters. The cluster centers are used as the nodes of the graph, and edges are added between nodes within $H_\text{TD}$. The resulting graph facilitates natural stitching opportunities between states from different trajectories in the dataset.

\subsection{Temporal Efficiency Metric}
\begin{figure}[H]
    \centering
    \includegraphics[width=0.95\linewidth]{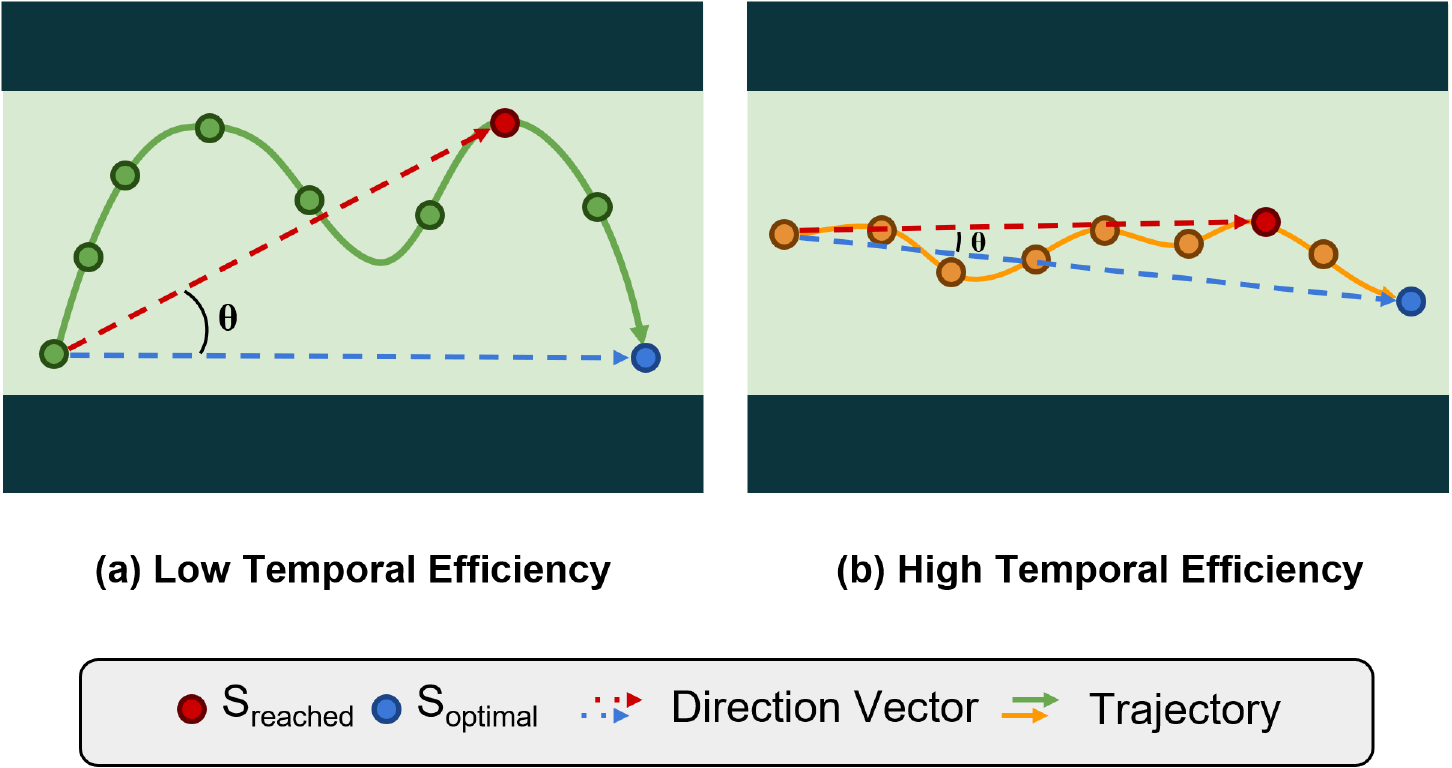}
    \caption{\textbf{Temporal Efficiency (TE).}  
    \textit{(left)} A transition with low TE, where the reached state diverges from the optimal direction. \textit{(right)} A transition with high TE, where the reached state aligns well with the optimal direction. TE measures the directional alignment between the actual and optimal transition vectors over a fixed temporal distance.}
    \label{fig:te}
\end{figure}

Using all states in the dataset for graph construction may not always be effective. In particular, when suboptimal trajectories are included, low-quality states can negatively impact stitching ability or final task performance. To address this, we introduce the Temporal Efficiency (TE) metric to selectively choose states for graph construction. TE measures how efficiently a state facilitates transitions by evaluating the similarity between states reached within the same trajectory over a fixed temporal distance. (Figure~\ref{fig:te})

Precisely, given a state \( s_{\text{cur}} \), we define the optimal future state \( s_{\text{opt}} \) as the state after a temporal distance of \( H_{\text{TD}} \) in the trajectory, and the actual reached state \( s_{\text{reached}} \) as the state observed \( H_{\text{TD}} \) steps ahead:
\begin{equation}
\mathcal{F}(s_t, d) = 
s_{\min \{ k \mid k > t,\ \|\psi(s_t) - \psi(s_k)\|_2 \geq d \}} 
\label{eq:opt_future_state_fn}
\end{equation}
\begin{equation}
s_{\text{opt}} = 
\mathcal{F}(s_{\text{cur}}, H_{\text{TD}}), \quad
s_{\text{reached}} = s_{\text{cur} + H_{\text{TD}}}
\end{equation}
If the trajectory was collected with optimal behavior, \( \psi(s_\text{reached}) \) and \( \psi(s_\text{opt}) \) are expected to be similar, whereas suboptimal trajectories often lead to larger discrepancies. 
We define the TE as the cosine similarity between the directional vectors, capturing the alignment between the optimal and actual transitions from the current state:
\begin{equation}
\theta_{\text{TE}} 
= \mathrm{cos}\left( \psi(s_{\text{opt}}) - \psi(s_{\text{cur}}),\ \psi(s_{\text{reached}}) - \psi(s_{\text{cur}}) \right)
\end{equation}
To ensure the quality of the constructed graph, we utilize only states with high TE values (e.g., \( \theta_{\text{TE}} \geq 0.9 \)) for graph construction (Section~\ref{sec:td_aware_graph_construction}), ensuring that only efficient transitions are incorporated. Through experiments, TE not only reduces computational overhead during the graph construction but also enhances overall task performance by improving the quality of generated nodes. The complete graph construction process, including the application of the TE metric, is shown in Appendix~\ref{app:graph_construction}.

\subsection{Low-Level Agent Training}
We construct a TD-aware dataset from an unlabeled (reward-free) dataset by applying a pretrained Temporal Distance Representation (TDR). We discuss two key components of low-level agent training: value function learning (Section~\ref{sec:value_function_learning}) and subgoal-conditioned policy learning (Section~\ref{sec:subgoal_conditioned_policy_learning}).

\subsubsection{Value Function Learning} 
\label{sec:value_function_learning}
We utilize the TD-aware intrinsic reward proposed in prior work for value function learning~\cite{park2024foundation}. The reward function is defined as:
\begin{equation}
r^\text{dir}(s, s', \vec{h}_\text{dir}) = 
\langle \psi(s') - \psi(s), \vec{h}_\text{dir} \rangle
\end{equation}
Given this reward, we optimize IQL~\cite{kostrikov2022offline, park2023hiql} based critic objectives conditioned on \( \vec{h}_\text{dir} \):
\begin{equation}
\mathcal{L}_{Q} =
\mathbb{E}_{\mathcal{D}}
        \Big[\big( r^\text{dir} + \gamma\,V(s', \vec{h}_\text{dir}) - Q(s,a, \vec{h}_\text{dir}) \big)^2 \Big]
\end{equation}
\begin{equation}
\mathcal{L}_{V} = 
\mathbb{E}_{\mathcal{D}}
        \Big[ \ell_\tau^2 \big( \bar{Q}(s,a, \vec{h}_\text{dir}) - V(s, \vec{h}_\text{dir}) \big) \Big]
\end{equation}
where $\bar{Q}$ is the target Q-function, $\ell_\tau^2(\cdot)$ is the expectile loss (Eq.~\eqref{eq:expectile}), and $\vec{h}_\text{dir}\sim\mathrm{Unif}(\mathbb{S}^{d-1})$ is a unit vector sampled uniformly from the unit sphere. Here, $\mathbb{S}^{d-1} := \{\, v \in \mathbb{R}^d \mid \|v\| = 1 \,\}$, and \( d \) is the dimension of the Temporal Distance Representation (TDR). This intrinsic reward provides a dense learning signal by measuring the directional alignment via the inner product with \( \vec{h}_\text{dir} \), thereby improving value prediction.

\begin{figure}[t]
    \centering
    \includegraphics[width=0.99\linewidth]{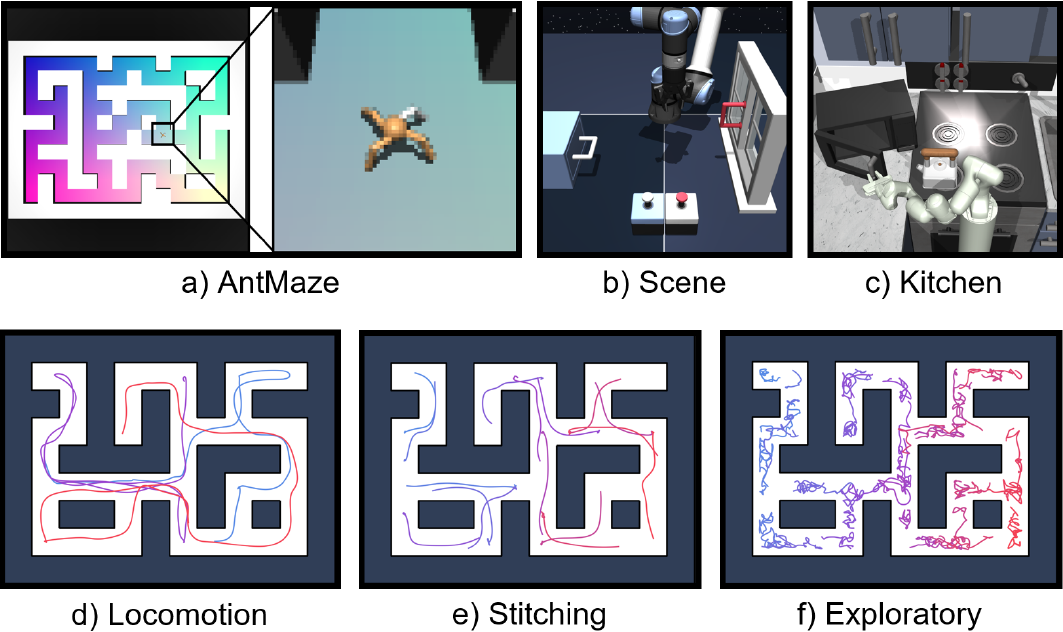}
    \caption{\textbf{Environments and Dataset Types.}
    \textit{(a)}, \textit{(b)}, and \textit{(c)} show representative environments used in offline goal-conditioned benchmarks. \textit{(d)}, \textit{(e)}, and \textit{(f)} illustrate three different dataset types in AntMaze~\cite{gupta2020relay, park2025ogbench}.}
\end{figure}

\subsubsection{Subgoal-Conditioned Policy Learning} 
\label{sec:subgoal_conditioned_policy_learning}
Previous offline hierarchical reinforcement learning (HRL) methods typically adopt step-based sampling~\cite{ajay2021opal, shin2023guide, li2024boosting}, where a subgoal is selected a fixed number of steps ahead (e.g., \( c \) steps) from the current state within the same trajectory~\cite{park2023hiql, park2024foundation}. However, a fixed \( c \)-step horizon may not align with the temporal distance separation between graph nodes defined by \( H_{\text{TD}} \), leading to inconsistency between training and task execution. To resolve this, we adopt a TD-aware subgoal sampling strategy that selects the state after a fixed temporal distance \( H_{\text{TD}} \) within the same trajectory. This corresponds to \( s_{\text{sub}} = \mathcal{F}(s_t, H_{\text{TD}}) \), as defined in Eq.~\eqref{eq:opt_future_state_fn}, and ensures that the subgoal horizon used for policy learning aligns with the one used during task execution. 

Once the subgoal is selected, its representation also plays a critical role in policy generalization. Previous studies have shown that leveraging normalized subgoal representations, rather than absolute ones, can improve downstream task performance~\cite{park2023hiql, park2024foundation}. To enhance generalization, we represent subgoals as a relative direction vector, where 
\( \vec{h}_{\mathrm{dir}} = \operatorname{dir}(\psi(s_t), \psi(s_{\text{sub}})) \):
\begin{equation}
\label{eq:directional_subgoal}
\operatorname{dir}(\psi(s_{t}),\psi(s_{\text{sub}})) = 
\frac{\psi(s_{\text{sub}}) - \psi(s_{t})}{\|\psi(s_{\text{sub}}) - \psi(s_{t})\|}
\end{equation}
To train the low-level policy, we employ deep deterministic policy gradient and behavioral cloning (DDPG+BC)~\cite{lillicrap2016continuous, fujimoto2021minimalist, park2024value} objective, which directly maximizes Q-values while constraining the learned policy to stay close to the behavior policy. The policy extraction objective, denoted as \( \mathcal{L}_{\text{DDPG+BC}} \), is defined as:
\begin{equation}
\mathbb{E}_{\mathcal{D}}
\Bigl[Q\!\bigl(s_{t},\,\mu^\pi(s_{t},\vec{h}_\text{dir}),\,\vec{h}_\text{dir}\bigr) + \alpha \log \pi\!\bigl(a \mid s_{t},\vec{h}_\text{dir}\bigr)\Bigr]
\end{equation}
where \( \mu^\pi \) denotes a deterministic policy that outputs the mean action, \( \alpha \) is a hyperparameter controlling the strength of the BC regularizer. 

\begin{algorithm}[H]
    \caption{Task Planning and Execution}
    \label{algo:task_planning_execution}
    \begin{algorithmic}
        \STATE \textbf{Input:} Graph $\mathcal{H}_{\text{graph}}=(\mathcal{V}, \mathcal{E})$, Low-Level Policy $\pi_{\text{low}}$, TDR $\psi$, Temporal Distance Threshold $H_{\text{TD}}$, Goal $g_{\text{final}}$

        \vspace{10pt} 
        
        \STATE \textbf{// Precompute Shortest Distance from Nodes to Goal}
        \STATE // $\mathrm{Dijkstra}(\psi(g_{\text{final}}), \mathcal{H}_{\text{graph}})=\{\,\mathrm{dist}_v\mid v\in\mathcal V\}$
        \STATE $\mathrm{Dists} \gets \mathrm{Dijkstra}(\psi(g_{\text{final}}), \mathcal{H}_{\text{graph}})$
        
        \vspace{10pt} 
        
        \STATE \textbf{// Task Planning \& Execution}
        \STATE Set current state: $s_{\text{cur}} =  \mathrm{env.reset}() $
        \STATE Compute TDR embedding: $h_{\text{cur}} = \psi(s_{\text{cur}})$
        
        \WHILE{$s_{\text{cur}} \neq g_{\text{final}}$}
            \STATE // Identify reachable nodes within $H_{\text{TD}}$ distance
            \STATE $\mathcal{V}_{\text{near}} \gets \{ v \in \mathcal{V} \mid \|h_{\text{cur}}-v\|_2 \leq H_{\text{TD}} \}$
            
            \vspace{10pt} 
            
            \STATE // Select subgoal with shortest distance to goal
            \STATE $v_{\text{subgoal}} = \arg\min_{v \in \mathcal{V}_{\text{near}}} \bigl(\mathrm{Dists}[v] + \|h_{\text{cur}}-v\|_2\bigr)$

            \vspace{10pt} 
            
            \STATE // Execute Low-Level Policy
            \STATE $\vec{h}_\text{dir} = \mathrm{dir}(h_{\text{cur}}, v_{\text{subgoal}})$ // Eq.~\eqref{eq:directional_subgoal} 
            \STATE $a_{\text{cur}} \sim \pi_{\text{low}}(s_{\text{cur}}, \vec{h}_\text{dir})$
            \STATE $s_{\text{cur}} = \mathrm{env.step}(a_{\text{cur}}) $
            \STATE $h_{\text{cur}} = \psi(s_{\text{cur}})$
        \ENDWHILE
    \end{algorithmic}
\end{algorithm}

\subsection{Task Planning and Execution}
GAS utilizes the constructed graph to support task planning and subgoal selection during the task execution of the learned low-level policy. The framework does not require high-level policy learning for subgoal generation. At the beginning of each episode, Dijkstra’s algorithm~\cite{dijkstra1959note} is used to precompute the shortest distances from all graph nodes to the final goal. Among the reachable nodes within temporal distance \( H_{\text{TD}} \), the agent selects the subgoal with the precomputed shortest distance. The low-level policy then predicts primitive actions to reach the selected subgoal. By removing the need for high-level policy learning, this method improves the overall efficiency and stability of the framework while achieving high performance in environments requiring stitching ability. (Algorithm~\ref{algo:task_planning_execution})

\begin{table*}[t]
\caption{Evaluating GAS on state-based offline goal-conditioned benchmarks.}
\label{tab:eval_gas_state} 
\centering
\resizebox{0.99\textwidth}{!}{
\begin{tabular}{l l r r r r r r r r}
\toprule
\textbf{Dataset Type} & \textbf{Dataset} & \textbf{GCBC} & \textbf{GCIQL} & \textbf{QRL} & \textbf{CRL} & \textbf{HGCBC} & \textbf{HHILP} & \textbf{HIQL} & \textbf{GAS (ours)} \\
\midrule
\multirow{3}{*}{Locomotion}
 & antmaze-medium-navigate & 33.1\;{\scriptsize$\pm$ 5.6} & 74.6\;{\scriptsize$\pm$ 4.8} & 81.9\;{\scriptsize$\pm$ 8.2} & \textbf{95.3}\;{\scriptsize$\pm$ 1.0} & 58.1\;{\scriptsize$\pm$ 5.5} & \textbf{96.3}\;{\scriptsize$\pm$ 0.4} & \textbf{95.3}\;{\scriptsize$\pm$ 1.3} & \textbf{96.3}\;{\scriptsize$\pm$ 1.3}\\
 & antmaze-large-navigate  & 23.4\;{\scriptsize$\pm$ 3.2} & 32.6\;{\scriptsize$\pm$ 4.7} & 74.9\;{\scriptsize$\pm$ 4.4} & 85.5\;{\scriptsize$\pm$ 5.3} & 44.3\;{\scriptsize$\pm$ 4.1} & 86.8\;{\scriptsize$\pm$ 3.6} & \textbf{89.9}\;{\scriptsize$\pm$ 2.2} & \textbf{93.2}\;{\scriptsize$\pm$ 0.5}\\
 & antmaze-giant-navigate  & 0.0\;{\scriptsize$\pm$ 0.0}  & 0.1\;{\scriptsize$\pm$ 0.4}  & 14.3\;{\scriptsize$\pm$ 3.6} & 15.0\;{\scriptsize$\pm$ 5.7} & 7.2\;{\scriptsize$\pm$ 1.7}  & 53.1\;{\scriptsize$\pm$ 2.6} & 67.3\;{\scriptsize$\pm$ 5.5} & \textbf{77.6}\;{\scriptsize$\pm$ 2.9}\\
\midrule
\multirow{3}{*}{Stitching}
 & antmaze-medium-stitch   & 43.2\;{\scriptsize$\pm$ 7.7} & 26.6\;{\scriptsize$\pm$ 6.8} & 67.0\;{\scriptsize$\pm$10.6} & 57.0\;{\scriptsize$\pm$ 7.9} & 65.9\;{\scriptsize$\pm$ 5.7} & \textbf{96.0}\;{\scriptsize$\pm$ 1.2} & 92.0\;{\scriptsize$\pm$ 2.8} & \textbf{98.1}\;{\scriptsize$\pm$ 1.2}\\
 & antmaze-large-stitch    & 2.3\;{\scriptsize$\pm$ 3.6}  & 9.6\;{\scriptsize$\pm$ 3.1}  & 20.2\;{\scriptsize$\pm$ 1.7} & 14.4\;{\scriptsize$\pm$ 5.9} & 10.7\;{\scriptsize$\pm$ 5.8} & 34.1\;{\scriptsize$\pm$ 3.0}   & 71.7\;{\scriptsize$\pm$ 4.8} & \textbf{96.3}\;{\scriptsize$\pm$ 0.9}\\
 & antmaze-giant-stitch    & 0.0\;{\scriptsize$\pm$ 0.0}  & 0.0\;{\scriptsize$\pm$ 0.0}  & 0.4\;{\scriptsize$\pm$ 0.3}  & 0.0\;{\scriptsize$\pm$ 0.0}  & 0.0\;{\scriptsize$\pm$ 0.0}  & 0.0\;{\scriptsize$\pm$ 0.0}  & 1.0\;{\scriptsize$\pm$ 1.2}  & \textbf{88.3}\;{\scriptsize$\pm$ 3.6}\\
\midrule
\multirow{2}{*}{Exploratory}
 & antmaze-medium-explore  & 2.7\;{\scriptsize$\pm$ 2.8}  & 11.7\;{\scriptsize$\pm$ 1.3} & 1.4\;{\scriptsize$\pm$ 1.2}  & 1.0\;{\scriptsize$\pm$ 1.6}  & 15.0\;{\scriptsize$\pm$ 8.2} & 39.9\;{\scriptsize$\pm$ 7.4} & 32.2\;{\scriptsize$\pm$ 3.0}   & \textbf{98.1}\;{\scriptsize$\pm$ 0.4}\\
 & antmaze-large-explore   & 0.0\;{\scriptsize$\pm$ 0.0}  & 0.6\;{\scriptsize$\pm$ 0.5}  & 0.3\;{\scriptsize$\pm$ 1.0}   & 0.0\;{\scriptsize$\pm$ 0.0}  & 0.0\;{\scriptsize$\pm$ 0.0}  & 2.4\;{\scriptsize$\pm$ 1.9}  & 2.9\;{\scriptsize$\pm$ 4.3}  & \textbf{94.2}\;{\scriptsize$\pm$ 3.0}\\
\midrule
\multirow{2}{*}{Manipulation}
 & scene-play              & 5.4\;{\scriptsize$\pm$ 0.9}  & 50.4\;{\scriptsize$\pm$ 1.4} & 5.1\;{\scriptsize$\pm$ 1.7}  & 19.2\;{\scriptsize$\pm$ 3.0}   & 4.6\;{\scriptsize$\pm$ 1.3}  & 43.4\;{\scriptsize$\pm$ 5.2} & 40.0\;{\scriptsize$\pm$ 9.6} & \textbf{73.6}\;{\scriptsize$\pm$ 8.0}\\
 & kitchen-partial         & 69.5\;{\scriptsize$\pm$14.1} & 55.6\;{\scriptsize$\pm$17.5} & 61.9\;{\scriptsize$\pm$ 8.5} & 32.7\;{\scriptsize$\pm$11.7} & 71.1\;{\scriptsize$\pm$ 6.2} & 66.7\;{\scriptsize$\pm$ 9.0}   & 73.1\;{\scriptsize$\pm$ 2.4} & \textbf{87.3}\;{\scriptsize$\pm$ 8.8}\\
\bottomrule
\end{tabular}
} 
\end{table*}

\begin{table*}[t]
\caption{Evaluating GAS on pixel-based offline goal-conditioned benchmarks.}
\label{tab:eval_gas_pixel}
\centering
\resizebox{0.86\textwidth}{!}{
\begin{tabular}{l l r r r r r r}
\toprule
\textbf{Dataset Type} & \textbf{Dataset} & \textbf{GCIQL} 
& \textbf{QRL} & \textbf{CRL} & \textbf{HHILP} & \textbf{HIQL} & \textbf{GAS (ours)} \\
\midrule
\multirow{3}{*}{Locomotion} 
 & visual-antmaze-medium-navigate & 19.1\;{\scriptsize$\pm$ 1.6} & 0.0\;{\scriptsize$\pm$ 0.0} & \textbf{93.7}\;{\scriptsize$\pm$ 1.2} & \textbf{94.1}\;{\scriptsize$\pm$ 1.2} & \textbf{95.5}\;{\scriptsize$\pm$ 0.8} & \textbf{96.4}\;{\scriptsize$\pm$ 0.5}\\
 & visual-antmaze-large-navigate  & 4.6\;{\scriptsize$\pm$ 1.9}  & 0.0\;{\scriptsize$\pm$ 0.0} & 79.5\;{\scriptsize$\pm$ 7.5} & \textbf{85.6}\;{\scriptsize$\pm$ 2.5} & 80.0\;{\scriptsize$\pm$ 2.1} & \textbf{87.0}\;{\scriptsize$\pm$ 1.2}\\
 & visual-antmaze-giant-navigate  & 1.5\;{\scriptsize$\pm$ 0.8}  & 0.2\;{\scriptsize$\pm$ 0.8} & 43.4\;{\scriptsize$\pm$ 5.9} & 42.4\;{\scriptsize$\pm$ 1.9} & 34.1\;{\scriptsize$\pm$ 14.0} & \textbf{59.0}\;{\scriptsize$\pm$ 2.1}\\
\midrule
\multirow{3}{*}{Stitching} 
 & visual-antmaze-medium-stitch   & 4.2\;{\scriptsize$\pm$ 1.6}  & 0.0\;{\scriptsize$\pm$ 0.0} & 68.0\;{\scriptsize$\pm$ 8.3} & \textbf{92.4}\;{\scriptsize$\pm$ 1.2} & \textbf{90.4}\;{\scriptsize$\pm$ 4.1} & \textbf{90.0}\;{\scriptsize$\pm$ 3.0}\\
 & visual-antmaze-large-stitch    & 0.2\;{\scriptsize$\pm$ 0.3}  & 0.1\;{\scriptsize$\pm$ 0.5} & 14.7\;{\scriptsize$\pm$ 7.1} & 33.8\;{\scriptsize$\pm$ 1.2} & 38.5\;{\scriptsize$\pm$ 5.7} & \textbf{75.2}\;{\scriptsize$\pm$ 4.4}\\
 & visual-antmaze-giant-stitch    & 0.0\;{\scriptsize$\pm$ 0.0}  & 0.2\;{\scriptsize$\pm$ 0.6} & 0.0\;{\scriptsize$\pm$ 0.0} & 3.6\;{\scriptsize$\pm$ 1.3} & 0.9\;{\scriptsize$\pm$ 1.1}  & \textbf{55.8}\;{\scriptsize$\pm$ 3.5}\\
\midrule
\multirow{2}{*}{Exploratory} 
 & visual-antmaze-medium-explore  & 0.0\;{\scriptsize$\pm$ 0.0}  & 0.1\;{\scriptsize$\pm$ 0.3} & 0.0\;{\scriptsize$\pm$ 0.0}  & 0.0\;{\scriptsize$\pm$ 0.0} & 0.9\;{\scriptsize$\pm$ 1.4}  & \textbf{65.9}\;{\scriptsize$\pm$ 6.8}\\
 & visual-antmaze-large-explore   & 0.0\;{\scriptsize$\pm$ 0.0}  & 0.0\;{\scriptsize$\pm$ 0.0} & 0.0\;{\scriptsize$\pm$ 0.0}  & 0.0\;{\scriptsize$\pm$ 0.0} & 0.0\;{\scriptsize$\pm$ 0.0}  & \textbf{15.1}\;{\scriptsize$\pm$ 6.8}\\
\midrule
\multirow{1}{*}{Manipulation} 
 & visual-scene-play              & 10.6\;{\scriptsize$\pm$ 2.7} & 13.5\;{\scriptsize$\pm$ 2.8} & 8.4\;{\scriptsize$\pm$ 0.9}  & 35.6\;{\scriptsize$\pm$ 4.9} & 47.9\;{\scriptsize$\pm$ 3.9} & \textbf{54.4}\;{\scriptsize$\pm$ 6.2}\\
\bottomrule
\end{tabular}
}
\end{table*}

\section{Experiments}
We evaluate GAS on OGBench~\cite{park2025ogbench} and D4RL~\cite{fu2020d4rl}, spanning diverse dataset types. We compare its performance against offline goal-conditioned and hierarchical baselines. For each dataset, we report the average normalized return across five test-time goals, except for kitchen, which uses a single fixed goal. Each goal is evaluated with 50 rollouts, and results are averaged over 4 random seeds. Bold numbers indicate results that are at least 95\% of the best-performing method in each row. Details of the datasets and baselines are provided in Appendices~\ref{app:environments_and_datasets},~\ref{app:implementation_details}.

\subsection{Main Results} 
We first present the evaluation results on numeric state environments in Table~\ref{tab:eval_gas_state}. Overall, our GAS achieves the best performance across all tasks. Among the baselines, HIQL performs well on the relatively easier navigation dataset, achieving a score of 67.3 in antmaze-giant-navigate. However, its performance dramatically drops to 1.0 in antmaze-giant-stitch, revealing its difficulty in trajectory stitching. In contrast, GAS significantly improves performance in the same task, achieving 88.3 by explicitly performing trajectory stitching through TD-aware Graph Construction. Furthermore, while baseline models struggle with low-quality datasets, failing to learn meaningful policies, GAS maintains strong performance. For example, in antmaze-large-explore, GAS achieves 94.2, whereas HHILP and HIQL achieve only 2.4 and 2.9, respectively.

Table~\ref{tab:eval_gas_pixel} presents the performance results in visual state environments. GAS outperforms the baselines in most cases. Notably, it is the only model that achieves meaningful performance in challenging tasks such as visual-antmaze-giant-stitch and visual-antmaze-large-explore. However, compared to numeric state environments, the overall performance is lower, which can be attributed to a lack of high-dimensional representation learning for visual input. In contrast, CRL shows improved performance in visual-antmaze-giant-navigate compared to its numeric counterpart, possibly due to its contrastive learning objective~\cite{eysenbach2022contrastive, zheng2024stabilizing}. To improve performance in visual state environments, we plan to enhance GAS by incorporating advanced representation learning techniques for visual feature extraction.

\subsection{Ablation Study}
\begingroup
\setlength{\tabcolsep}{9pt}    
\begin{table*}[t]
\caption{\textbf{The Temporal Efficiency (TE) metric improves task performance while reducing computational overhead.} This table compares graphs constructed using all states and those built from states filtered by the TE metric. The TE-filtered States (\%) column reports the percentage of states used for clustering, which significantly lowers graph construction overhead while leading to higher-quality nodes and improved task performance. As shown in the \# Nodes in Graph column, the number of nodes is consistently fewer than 1\% of the total dataset states across all tasks in this table.}
\label{tab:te_metric_ablation}
\centering
\resizebox{0.99\textwidth}{!}{
\begin{tabular}{lcccccccc}
\toprule
\multirow[c]{2}{*}[-0.45ex]{\textbf{Dataset}} &
\multirow[c]{2}{*}[-0.45ex]{\textbf{\# States in Dataset}} &
\multicolumn{2}{c}{\textbf{TE‑Filtered States (\%)}} &
\multicolumn{2}{c}{\textbf{\# Nodes in Graph}} &
\multicolumn{3}{c}{\textbf{Normalized Return}} \\
\cmidrule(lr){3-4} \cmidrule(lr){5-6} \cmidrule(lr){7-9}
&
& \textbf{All States} & \textbf{Ours}
& \textbf{All States} & \textbf{Ours}
& \textbf{All States} & \textbf{Ours} & $\boldsymbol{\Delta\uparrow}$ \\
\midrule
antmaze‑giant‑navigate & 1M & 100 & 6 & 2092 & 978 & 63.4\;{\scriptsize$\pm$ 3.7} & \textbf{77.6}\;{\scriptsize$\pm$ 2.9} & +14.2 \\
antmaze‑giant‑stitch   & 1M & 100 & 8 & 3490 & 1966 & 75.3\;{\scriptsize$\pm$ 5.7} & \textbf{88.3}\;{\scriptsize$\pm$ 3.6} & +13.0 \\
antmaze‑large‑explore  & 5M & 100 & 2 & 6213 & 2499 & 75.4\;{\scriptsize$\pm$ 4.3} & \textbf{94.2}\;{\scriptsize$\pm$ 3.0} & +18.8 \\
scene‑play             & 1M & 100 & 6 & 2809 & 725  & 63.5\;{\scriptsize$\pm$ 5.5} & \textbf{73.6}\;{\scriptsize$\pm$ 8.0} & +10.1 \\
\bottomrule
\end{tabular}}
\end{table*}
\endgroup

\begin{figure*}[t]
\centering
\includegraphics[width=0.9\textwidth]{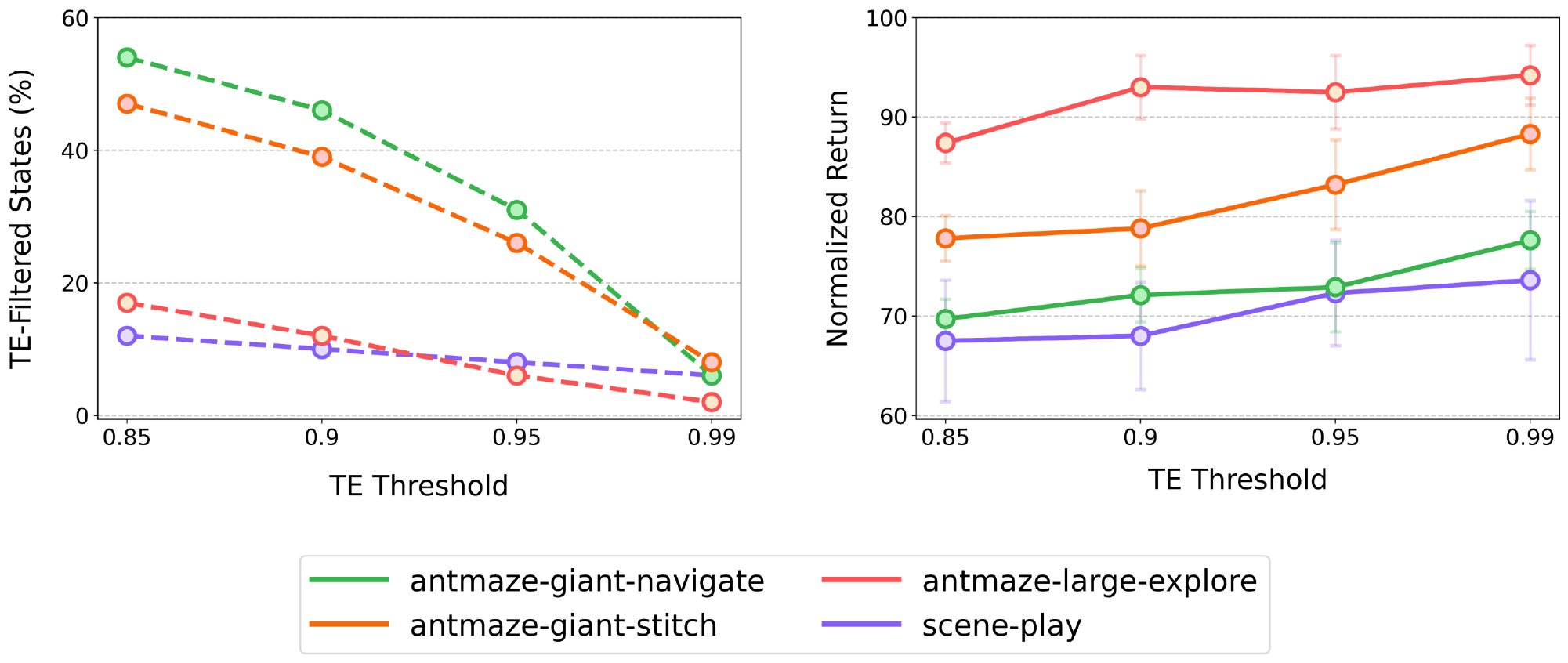}
\caption{\textbf{Impact of the Temporal Efficiency (TE) threshold $\theta_{TE}^{\text{thresh}}$ on graph construction and task performance.}
\textit{(left)} Percentage of states retained for clustering, relative to the total number of states in the dataset.
\textit{(right)} Normalized return achieved under each $\theta_{TE}^{\text{thresh}}$. These results show that, with an appropriately chosen $\theta_{TE}^{\text{thresh}}$, TE filtering significantly reduces graph construction overhead while preserving or improving task performance.}
\label{fig:te_threshold_ablation}
\end{figure*}

\subsubsection{Temporal Efficiency Filtering} 
We investigate the impact of filtering inefficient transition states using the Temporal Efficiency (TE) metric and constructing a graph exclusively with high-quality states. Using all states for graph construction significantly increases the computational cost of clustering, particularly in large-scale datasets. Moreover, in datasets with many noisy transitions, the resulting graph may contain inefficient nodes that degrade subgoal selection quality during task planning. Table~\ref{tab:te_metric_ablation} compares the performance of GAS with TE filtering to a variant that constructs the graph from all states without filtering. Experimental results show that constructing the graph from a small number of high-quality states, selected via the TE metric, not only reduces graph construction overhead but also improves task performance. The effect is particularly pronounced in the antmaze-large-explore, which contains many suboptimal trajectories. This indicates that GAS enables more refined subgoal selection by filtering out inefficient transition states from low-quality datasets.

\subsubsection{Temporal Efficiency Threshold} 
We further analyze the impact of the Temporal Efficiency (TE) threshold $\theta_{TE}^{\text{thresh}}$ on both the proportion of retained states and task performance, as shown in Figure~\ref{fig:te_threshold_ablation}. 
For example, when the TE threshold is set to 0.99, approximately 2\% to 8\% of the states are retained, depending on the dataset. This clearly demonstrates that filtering low-efficiency states in advance not only reduces graph construction overhead but also helps preserve or improve task performance. Note that the TE-Filtered States (\%) column refers exclusively to the graph construction process. During low-level policy training, the entire dataset was utilized. Additionally, for all tested values of $\theta_{TE}^{\text{thresh}}$ between 0.9 and 0.99, the number of nodes remains below 1\% of the total dataset states in all numeric state environments, which also helps reduce the computational cost during task planning. Additional statistics on the number of graph nodes for all datasets are provided in Appendix~\ref{app:environments_and_datasets}.

\subsubsection{Graph Node Selection Method}
\begin{figure}[H]
\centering
\includegraphics[width=0.95\linewidth]{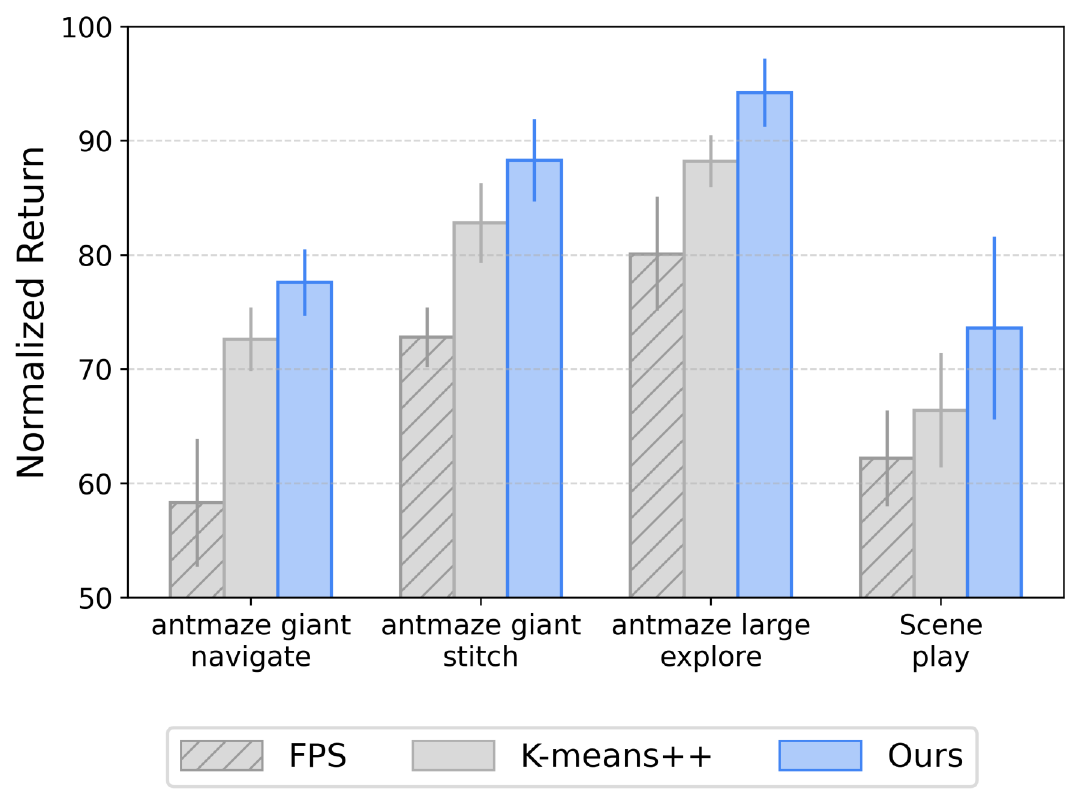}
\caption{\textbf{Comparison of graph node selection methods in the Temporal Distance Representation (TDR) space.} Our method explicitly enforces uniform separation in temporal distance between nodes, leading to consistently better task performance.}
\label{fig:td_aware_clustering_ablation}
\end{figure}

In this experiment, we compare different graph node selection methods, as presented in Figure~\ref{fig:td_aware_clustering_ablation}. Specifically, we evaluate two commonly used node selection algorithms: Farthest Point Sampling (FPS)~\cite{Eldar1994}, widely adopted in prior graph-based reinforcement learning (RL) studies~\cite{kim2021landmark, lee2022dhrl, kim2023imitating, park2024ngte}, and K-Means++~\cite{arthur2007kmeanspp}, and compare them to our proposed method. Both baseline methods construct graphs in the Temporal Distance Representation (TDR) space using latent states. Additionally, to ensure a fair comparison, all methods use the same number of graph nodes as GAS. Experimental results show that GAS consistently outperforms both FPS and K-Means++ across all tasks. This performance gain stems from enforcing uniform separation in temporal distance between nodes, which enhances graph connectivity and improves subgoal selection during task planning.

\subsubsection{Low-Level Subgoal Sampling Strategy}
In this experiment, we compare three subgoal sampling strategies used during low-level policy training, as shown in Table~\ref{tab:subgoal_sampling_ablation}. The first is random direction sampling, where directions are sampled from a uniform distribution. The second is step-based sampling, where a subgoal is selected a fixed number of steps ahead from the current state within the same trajectory. The last is our subgoal sampling strategy, which selects a subgoal located at a fixed temporal distance $H_{\text{TD}}$ within the same trajectory. Experimental results show that GAS consistently outperforms both baselines across all tasks. This performance gain stems from the alignment between the sampling horizon and the temporal distance between graph nodes, ensuring temporal consistency between training and task execution. The effect becomes more pronounced in environments with a large $H_{\text{TD}}$. Note that these subgoal sampling strategies are used only during low-level policy training, while the value function is trained independently using randomly sampled directions. 

\subsubsection{Temporal Distance Threshold}
The temporal distance threshold $H_{\text{TD}}$ serves as a key hyperparameter that brings structural consistency to the GAS framework. It is used across multiple components, including Temporal Efficiency (TE) filtering, TD-aware clustering, and edge connection during graph construction. In addition, $H_{\text{TD}}$ is applied to subgoal sampling during low-level policy training and subgoal selection during task planning. As shown in Table~\ref{tab:temporal_distance_threshold_ablation}, we observe that the effect of $H_{\text{TD}}$ on performance varies across tasks. This indicates that selecting an appropriate $H_{\text{TD}}$ per environment can improve graph quality and overall task performance.

\begin{table}[H]
\centering
\caption{\textbf{Comparison of subgoal sampling strategies during low-level policy training.} Our subgoal sampling strategy aligns subgoal selection with the temporal distance $H_{\text{TD}}$ between graph nodes, maintaining temporal consistency and improving task performance, particularly for large $H_{\text{TD}}$.}
\label{tab:subgoal_sampling_ablation}
\begin{adjustbox}{width=\columnwidth}
\begin{tabular}{lcccc}
\toprule
\textbf{Dataset} & \textbf{$H_{\text{TD}}$} &
\makecell{\textbf{Random Direction}\\[-2pt]\textbf{Sampling}} &
\makecell{\textbf{Step-based}\\[-2pt]\textbf{Sampling}} &
\textbf{Ours} \\  
\midrule
antmaze‑giant‑navigate &  8 & 23.3\;{\scriptsize$\pm$\,4.0}  & \textbf{74.9}\;{\scriptsize$\pm$\,5.1} & \textbf{77.6}\;{\scriptsize$\pm$\,2.9} \\
antmaze‑giant‑stitch   &  8 &  5.3\;{\scriptsize$\pm$\,1.8}  & \textbf{86.0}\;{\scriptsize$\pm$\,2.7} & \textbf{88.3}\;{\scriptsize$\pm$\,3.6} \\
antmaze‑large‑explore  &  8 & 83.6\;{\scriptsize$\pm$\,6.1}  & \textbf{91.4}\;{\scriptsize$\pm$\,2.8} & \textbf{94.2}\;{\scriptsize$\pm$\,3.0} \\
scene‑play             & 48 & 35.8\;{\scriptsize$\pm$\,5.5}  & 17.6\;{\scriptsize$\pm$\,4.7} & \textbf{73.6}\;{\scriptsize$\pm$\,8.0} \\
\midrule
\textbf{Average}       &  & 37.0 & 67.5 & \textbf{83.4} \\
\bottomrule
\end{tabular}
\end{adjustbox}
\end{table}

\begin{table}[H]
\centering
\caption{\textbf{Impact of the temporal distance threshold $H_{\text{TD}}$ on graph construction and task performance.}
Selecting an appropriate $H_{\text{TD}}$ per task can improve graph quality and overall task performance.}
\label{tab:temporal_distance_threshold_ablation}
\begin{adjustbox}{width=\columnwidth}
\begin{tabular}{lccc}
\toprule
\textbf{Dataset} & $H_{\text{TD}}$ & \# Nodes in Graph & Normalized Return \\
\midrule
\multirow{4}{*}{antmaze-giant-navigate}
  & 4            & 4286 & \textbf{74.4}\;{\scriptsize$\pm$\,3.1} \\
  & 8 (ours)     & 978  & \textbf{77.6}\;{\scriptsize$\pm$\,2.9} \\
  & 12           & 431  & 70.8\;{\scriptsize$\pm$\,1.5} \\
  & 16           & 268  & 64.1\;{\scriptsize$\pm$\,2.2} \\
\midrule
\multirow{4}{*}{antmaze-giant-stitch}
  & 4            & 12901 & 80.7\;{\scriptsize$\pm$\,3.1} \\
  & 8 (ours)     & 1966  & \textbf{88.3}\;{\scriptsize$\pm$\,3.6} \\
  & 12           & 688   & 80.2\;{\scriptsize$\pm$\,4.0} \\
  & 16           & 375   & 68.5\;{\scriptsize$\pm$\,2.2} \\
\midrule
\multirow{4}{*}{antmaze-large-explore}
  & 4            & 15143 & \textbf{98.2}\;{\scriptsize$\pm$\,1.3} \\
  & 8 (ours)     & 2499  & \textbf{94.2}\;{\scriptsize$\pm$\,3.0} \\
  & 12           & 1126  & 87.6\;{\scriptsize$\pm$\,7.6} \\
  & 16           & 679   & 90.2\;{\scriptsize$\pm$\,3.9} \\
\midrule
\multirow{4}{*}{scene-play}
  & 44           & 855  & \textbf{71.5}\;{\scriptsize$\pm$\,10.2} \\
  & 48 (ours)    & 725  & \textbf{73.6}\;{\scriptsize$\pm$\,8.0} \\
  & 52           & 600  & \textbf{70.7}\;{\scriptsize$\pm$\,0.9} \\
  & 56           & 478  & \textbf{72.5}\;{\scriptsize$\pm$\,8.7} \\
\bottomrule
\end{tabular}
\end{adjustbox}
\end{table}

\section{Conclusion}
We proposed Graph-Assisted Stitching (GAS), a novel offline hierarchical reinforcement learning (HRL) framework that eliminates the need for explicit high-level policy learning. Instead, GAS formulates subgoal selection as a graph-based planning problem, improving long-horizon reasoning and trajectory stitching. To achieve this, GAS constructs a graph using Temporal Distance Representation (TDR) and clusters semantically similar states from different trajectory segments into unified graph nodes, enabling efficient transition stitching. To improve graph quality and reduce construction overhead, GAS filters out inefficient transitions by applying a Temporal Efficiency (TE) metric. Through extensive experiments on OGBench and D4RL benchmarks, GAS consistently outperforms prior methods across diverse offline goal-conditioned tasks. In particular, GAS achieves a score of 88.3 in antmaze-giant-stitch and 94.2 in antmaze-large-explore, dramatically surpassing the previous state-of-the-art scores of 1.0 and 2.9, respectively.

\section*{Impact Statement}
This paper presents work whose goal is to advance the field of Machine Learning. There are many potential societal consequences of our work, none which we feel must be specifically highlighted here.

\section*{Acknowledgement} This work was partly supported by the National Research Foundation of Korea (NRF) grant funded by the Korea government (MSIT) (No. RS-2024-00348376, High-Intelligence, High-Versatility, High-Adaptability Reinforcement Learning Methods for Real-World Composite Task Robots) and the Institute of Information Communications Technology Planning Evaluation (IITP) grant funded by the Korea government (MSIT) (No. RS-2024-00438686, Development of software reliability improvement technology through identification of abnormal open sources and automatic application of DevSecOps), (No. RS-2022-II221045, Self-directed Multi-Modal Intelligence for solving unknown open domain problems), (No. RS-2025-02218768, Accelerated Insight Reasoning via Continual Learning) and (No. RS-2019-II190421, Artificial Intelligence Graduate School Program).

\bibliography{icml2025}

\begin{thebibliography}{69}
\providecommand{\natexlab}[1]{#1}
\providecommand{\url}[1]{\texttt{#1}}
\expandafter\ifx\csname urlstyle\endcsname\relax
  \providecommand{\doi}[1]{doi: #1}\else
  \providecommand{\doi}{doi: \begingroup \urlstyle{rm}\Url}\fi

\bibitem[Ajay et~al.(2021)Ajay, Kumar, Agrawal, Levine, and Nachum]{ajay2021opal}
Ajay, A., Kumar, A., Agrawal, P., Levine, S., and Nachum, O.
\newblock Opal: Offline primitive discovery for accelerating offline reinforcement learning.
\newblock In \emph{International Conference on Learning Representations (ICLR)}, 2021.

\bibitem[Andrychowicz et~al.(2017)Andrychowicz, Wolski, Ray, Schneider, Fong, Welinder, McGrew, Tobin, Abbeel, and Zaremba]{andrychowicz2017hindsight}
Andrychowicz, M., Wolski, F., Ray, A., Schneider, J., Fong, R., Welinder, P., McGrew, B., Tobin, J., Abbeel, P., and Zaremba, W.
\newblock Hindsight experience replay.
\newblock In \emph{Neural Information Processing Systems (NeurIPS)}, 2017.

\bibitem[Arthur \& Vassilvitskii(2007)Arthur and Vassilvitskii]{arthur2007kmeanspp}
Arthur, D. and Vassilvitskii, S.
\newblock k-means++: The advantages of careful seeding.
\newblock In \emph{Symposium on Discrete Algorithms (SODA)}, 2007.

\bibitem[Ba et~al.(2016)Ba, Kiros, and Hinton]{layernorm}
Ba, J., Kiros, J.~R., and Hinton, G.~E.
\newblock Layer normalization.
\newblock \emph{ArXiv, abs/1607.06450}, 2016.

\bibitem[Bagaria et~al.(2020)]{bagaria2020skill}
Bagaria, A. et~al.
\newblock Skill discovery for exploration and planning using deep skill graphs.
\newblock In \emph{International Conference on Machine Learning (ICML)}, 2020.

\bibitem[Barto \& Mahadevan(2003)Barto and Mahadevan]{barto2003recent}
Barto, A.~G. and Mahadevan, S.
\newblock Recent advances in hierarchical reinforcement learning.
\newblock \emph{Discrete event dynamic systems}, 13:\penalty0 341--379, 2003.

\bibitem[Bradbury et~al.(2018)Bradbury, Frostig, Hawkins, Johnson, Leary, Maclaurin, Necula, Paszke, VanderPlas, Wanderman-Milne, and Zhang]{bradbury2018jax}
Bradbury, J., Frostig, R., Hawkins, P., Johnson, M.~J., Leary, C., Maclaurin, D., Necula, G., Paszke, A., VanderPlas, J., Wanderman-Milne, S., and Zhang, Q.
\newblock {JAX}: Composable transformations of {Python}+{NumPy} programs, 2018.
\newblock URL \url{https://github.com/google/jax}.

\bibitem[Brockman et~al.(2016)Brockman, Cheung, Pettersson, Schneider, Schulman, Tang, and Zaremba]{brockman2016gym}
Brockman, G., Cheung, V., Pettersson, L., Schneider, J., Schulman, J., Tang, J., and Zaremba, W.
\newblock Openai gym.
\newblock \emph{ArXiv, abs/1606.01540}, 2016.

\bibitem[Cao et~al.(2024)Cao, Yan, Lu, Tan, and Wang]{cao2024offline}
Cao, C., Yan, Z., Lu, R., Tan, J., and Wang, X.
\newblock Offline goal-conditioned reinforcement learning for safety-critical tasks with recovery policy.
\newblock In \emph{International Conference on Robotics and Automation (ICRA)}, 2024.

\bibitem[Dijkstra(1959)]{dijkstra1959note}
Dijkstra, E.~W.
\newblock A note on two problems in connexion with graphs.
\newblock \emph{Numerische Mathematik}, 1\penalty0 (1):\penalty0 269--271, 1959.

\bibitem[Eldar et~al.(1994)Eldar, Lindenbaum, Porat, and Zeevi]{Eldar1994}
Eldar, Y., Lindenbaum, M., Porat, M., and Zeevi, Y.~Y.
\newblock The farthest point strategy for progressive image sampling.
\newblock In \emph{International Conference on Pattern Recognition (ICPR)}, 1994.

\bibitem[Espeholt et~al.(2018)Espeholt, Soyer, Munos, Simonyan, Mnih, Ward, Doron, Firoiu, Harley, Dunning, Legg, and Kavukcuoglu]{espeholt2018impala}
Espeholt, L., Soyer, H., Munos, R., Simonyan, K., Mnih, V., Ward, T., Doron, Y., Firoiu, V., Harley, T., Dunning, I., Legg, S., and Kavukcuoglu, K.
\newblock Impala: Scalable distributed deep-rl with importance weighted actor-learner architectures.
\newblock In \emph{International Conference on Machine Learning (ICML)}, 2018.

\bibitem[Eysenbach et~al.(2019)Eysenbach, Salakhutdinov, and Levine]{eysenbach2019search}
Eysenbach, B., Salakhutdinov, R., and Levine, S.
\newblock Search on the replay buffer: Bridging planning and reinforcement learning.
\newblock In \emph{Neural Information Processing Systems (NeurIPS)}, 2019.

\bibitem[Eysenbach et~al.(2022)Eysenbach, Zhang, Levine, and Salakhutdinov]{eysenbach2022contrastive}
Eysenbach, B., Zhang, T., Levine, S., and Salakhutdinov, R.
\newblock Contrastive learning as goal-conditioned reinforcement learning.
\newblock In \emph{Neural Information Processing Systems (NeurIPS)}, 2022.

\bibitem[Faust et~al.(2018)Faust, Oslund, Ramirez, et~al.]{faust2018prmrl}
Faust, A., Oslund, K., Ramirez, O., et~al.
\newblock Prm-rl: Long-range robotic navigation tasks by combining reinforcement learning and sampling-based planning.
\newblock In \emph{International Conference on Robotics and Automation (ICRA)}, 2018.

\bibitem[Fu et~al.(2020)Fu, Kumar, Nachum, Tucker, and Levine]{fu2020d4rl}
Fu, J., Kumar, A., Nachum, O., Tucker, G., and Levine, S.
\newblock D4rl: Datasets for deep data-driven reinforcement learning.
\newblock \emph{ArXiv, abs/2004.07219}, 2020.

\bibitem[Fujimoto \& Gu(2021)Fujimoto and Gu]{fujimoto2021minimalist}
Fujimoto, S. and Gu, S.~S.
\newblock A minimalist approach to offline reinforcement learning.
\newblock In \emph{Neural Information Processing Systems (NeurIPS)}, 2021.

\bibitem[Fujimoto et~al.(2019)Fujimoto, Meger, and Precup]{fujimoto2019off}
Fujimoto, S., Meger, D., and Precup, D.
\newblock Off-policy deep reinforcement learning without exploration.
\newblock In \emph{International Conference on Machine Learning (ICML)}, 2019.

\bibitem[Ghosh et~al.(2021)Ghosh, Gupta, Reddy, Fu, Devin, Eysenbach, and Levine]{ghosh2021learning}
Ghosh, D., Gupta, A., Reddy, A., Fu, J., Devin, C., Eysenbach, B., and Levine, S.
\newblock Learning to reach goals via iterated supervised learning.
\newblock In \emph{International Conference on Learning Representations (ICLR)}, 2021.

\bibitem[Ghugare et~al.(2024)Ghugare, Lampert, Kipf, Anand, and Thomas]{ghugare2024temporal}
Ghugare, V., Lampert, C.~H., Kipf, T., Anand, A., and Thomas, V.
\newblock Closing the gap between td learning and supervised learning: A generalisation point of view.
\newblock In \emph{International Conference on Learning Representations (ICLR)}, 2024.

\bibitem[Gulcehre et~al.(2020)Gulcehre, Wang, Novikov, Smith, Shao, Szepesvari, Barreto, and Munos]{gulcehre2020rlunplugged}
Gulcehre, C., Wang, Z., Novikov, A., Smith, V., Shao, Y., Szepesvari, C., Barreto, A., and Munos, R.
\newblock Rl unplugged: Benchmarks for offline reinforcement learning.
\newblock In \emph{Neural Information Processing Systems (NeurIPS)}, 2020.

\bibitem[Gupta et~al.(2020)Gupta, Kumar, Lynch, Levine, and Hausman]{gupta2020relay}
Gupta, A., Kumar, V., Lynch, C., Levine, S., and Hausman, K.
\newblock Relay policy learning: Solving long-horizon tasks via imitation and reinforcement learning.
\newblock In \emph{Conference on Robot Learning (CoRL)}, 2020.

\bibitem[Haarnoja et~al.(2018{\natexlab{a}})Haarnoja, Zhou, Abbeel, and Levine]{haarnoja2018latent}
Haarnoja, T., Zhou, A., Abbeel, P., and Levine, S.
\newblock Latent space policies for hierarchical reinforcement learning.
\newblock In \emph{International Conference on Machine Learning (ICML)}, 2018{\natexlab{a}}.

\bibitem[Haarnoja et~al.(2018{\natexlab{b}})Haarnoja, Zhou, Abbeel, and Levine]{haarnoja2018sac}
Haarnoja, T., Zhou, A., Abbeel, P., and Levine, S.
\newblock Soft actor-critic: Off-policy maximum entropy deep reinforcement learning with a stochastic actor.
\newblock In \emph{International Conference on Machine Learning (ICML)}, 2018{\natexlab{b}}.

\bibitem[Hendrycks \& Gimpel(2016)Hendrycks and Gimpel]{gelu}
Hendrycks, D. and Gimpel, K.
\newblock Gaussian error linear units (gelus).
\newblock \emph{ArXiv, abs/1606.08415}, 2016.

\bibitem[Hoang et~al.(2021)Hoang, Sohn, Choi, Carvalho, and Lee]{hoang2021successor}
Hoang, C., Sohn, S., Choi, J., Carvalho, W., and Lee, H.
\newblock Successor feature landmarks for long-horizon goal-conditioned reinforcement learning.
\newblock In \emph{Neural Information Processing Systems (NeurIPS)}, 2021.

\bibitem[Jeong et~al.(2024)Jeong, Kim, Kim, Baek, Lee, Kim, and Ahn]{jeong2024prediction}
Jeong, H., Kim, D., Kim, D.~W., Baek, S., Lee, H.-C., Kim, Y., and Ahn, H.~J.
\newblock Prediction of intraoperative hypotension using deep learning models based on non-invasive monitoring devices.
\newblock \emph{Journal of Clinical Monitoring and Computing}, 38\penalty0 (6):\penalty0 1357--1365, 2024.

\bibitem[Kaelbling(1993)]{kaelbling1993learning}
Kaelbling, L.~P.
\newblock Learning to achieve goals.
\newblock In \emph{International Joint Conference on Artificial Intelligence (IJCAI)}, 1993.

\bibitem[Kim et~al.(2021)Kim, Seo, and Shin]{kim2021landmark}
Kim, J., Seo, Y., and Shin, J.
\newblock Landmark-guided subgoal generation in hierarchical reinforcement learning.
\newblock In \emph{Neural Information Processing Systems (NeurIPS)}, 2021.

\bibitem[Kim et~al.(2023)Kim, Seo, Ahn, Son, and Shin]{kim2023imitating}
Kim, J., Seo, Y., Ahn, S., Son, K., and Shin, J.
\newblock Imitating graph-based planning with goal-conditioned policies.
\newblock In \emph{International Conference on Learning Representations (ICLR)}, 2023.

\bibitem[Kim et~al.(2024)Kim, Choi, Matsunaga, and Kim]{kim2024ssd}
Kim, S., Choi, Y., Matsunaga, D.~E., and Kim, K.-E.
\newblock Stitching sub-trajectories with conditional diffusion model for goal-conditioned offline rl.
\newblock In \emph{Association for the Advancement of Artificial Intelligence (AAAI)}, 2024.

\bibitem[Kingma \& Ba(2015)Kingma and Ba]{kingma2015adam}
Kingma, D.~P. and Ba, J.
\newblock Adam: A method for stochastic optimization.
\newblock In \emph{International Conference on Learning Representations (ICLR)}, 2015.

\bibitem[Kostrikov et~al.(2021)Kostrikov, Yarats, and Fergus]{kostrikov2021image}
Kostrikov, I., Yarats, D., and Fergus, R.
\newblock Image augmentation is all you need: Regularizing deep reinforcement learning from pixels.
\newblock In \emph{International Conference on Learning Representations (ICLR)}, 2021.

\bibitem[Kostrikov et~al.(2022)Kostrikov, Nair, and Levine]{kostrikov2022offline}
Kostrikov, I., Nair, A., and Levine, S.
\newblock Offline reinforcement learning with implicit q-learning.
\newblock In \emph{International Conference on Learning Representations (ICLR)}, 2022.

\bibitem[Kumar et~al.(2019)Kumar, Fu, Soh, Tucker, and Levine]{kumar2019stabilizing}
Kumar, A., Fu, J., Soh, M., Tucker, G., and Levine, S.
\newblock Stabilizing off-policy q-learning via bootstrapping error reduction.
\newblock In \emph{Neural Information Processing Systems (NeurIPS)}, 2019.

\bibitem[Kumar et~al.(2020)Kumar, Zhou, Tucker, and Levine]{kumar2020conservative}
Kumar, A., Zhou, A., Tucker, G., and Levine, S.
\newblock Conservative q-learning for offline reinforcement learning.
\newblock In \emph{Neural Information Processing Systems (NeurIPS)}, 2020.

\bibitem[Kumar \& Todorov(2015)Kumar and Todorov]{kumar2015mujoco}
Kumar, V. and Todorov, E.
\newblock Mujoco haptix: A virtual reality system for hand manipulation.
\newblock In \emph{International Conference on Humanoid Robots (Humanoids)}, pp.\  657--663, 2015.

\bibitem[Lange et~al.(2012)Lange, Gabel, and Riedmiller]{lange2012batch}
Lange, S., Gabel, T., and Riedmiller, M.
\newblock Batch reinforcement learning.
\newblock In \emph{Reinforcement Learning: State-of-the-Art}, pp.\  45--73. Springer, 2012.

\bibitem[Lee et~al.(2022)Lee, Kim, Jang, and Kim]{lee2022dhrl}
Lee, S., Kim, J., Jang, I., and Kim, H.~J.
\newblock {DHRL}: A graph-based approach for long-horizon and sparse hierarchical reinforcement learning.
\newblock In \emph{Neural Information Processing Systems (NeurIPS)}, 2022.

\bibitem[Levine et~al.(2020)Levine, Kumar, Tucker, and Fu]{levine2020offline}
Levine, S., Kumar, A., Tucker, G., and Fu, J.
\newblock Offline reinforcement learning: Tutorial, review, and perspectives on open problems.
\newblock \emph{ArXiv, abs/2005.01643}, 2020.

\bibitem[Levy et~al.(2019)Levy, Konidaris, Platt, and Saenko]{levy2019learning}
Levy, A., Konidaris, G., Platt, R., and Saenko, K.
\newblock Learning multi-level hierarchies with hindsight.
\newblock In \emph{International Conference on Learning Representations (ICLR)}, 2019.

\bibitem[Li et~al.(2024{\natexlab{a}})Li, Shan, Zhu, Long, and Zhang]{li2024diffstitch}
Li, G., Shan, Y., Zhu, Z., Long, T., and Zhang, W.
\newblock Diffstitch: Boosting offline reinforcement learning with diffusion-based trajectory stitching.
\newblock In \emph{International Conference on Machine Learning (ICML)}, 2024{\natexlab{a}}.

\bibitem[Li et~al.(2024{\natexlab{b}})Li, Nie, Sun, Da, and Zhao]{li2024boosting}
Li, Z., Nie, F., Sun, Q., Da, F., and Zhao, H.
\newblock Boosting offline reinforcement learning for autonomous driving with hierarchical latent skills.
\newblock In \emph{International Conference on Robotics and Automation (ICRA)}, 2024{\natexlab{b}}.

\bibitem[Lillicrap et~al.(2016)Lillicrap, Hunt, Pritzel, Heess, Erez, Tassa, Silver, and Wierstra]{lillicrap2016continuous}
Lillicrap, T.~P., Hunt, J.~J., Pritzel, A., Heess, N., Erez, T., Tassa, Y., Silver, D., and Wierstra, D.
\newblock Continuous control with deep reinforcement learning.
\newblock In \emph{International Conference on Learning Representations (ICLR)}, 2016.

\bibitem[Lynch et~al.(2019)Lynch, Khansari, Xiao, Kumar, Tompson, Levine, and Sermanet]{lynch2019learning}
Lynch, C., Khansari, M., Xiao, T., Kumar, V., Tompson, J., Levine, S., and Sermanet, P.
\newblock Learning latent plans from play.
\newblock In \emph{Conference on Robot Learning (CoRL)}, 2019.

\bibitem[Ma et~al.(2022)Ma, Yan, Jayaraman, and Bastani]{ma2022gofar}
Ma, Y.~J., Yan, J., Jayaraman, D., and Bastani, O.
\newblock How far i’ll go: Offline goal-conditioned reinforcement learning via f-advantage regression.
\newblock In \emph{Neural Information Processing Systems (NeurIPS)}, 2022.

\bibitem[Nachum et~al.(2018)Nachum, Gu, Lee, and Levine]{nachum2018data}
Nachum, O., Gu, S., Lee, H., and Levine, S.
\newblock Data-efficient hierarchical reinforcement learning.
\newblock In \emph{Neural Information Processing Systems (NeurIPS)}, 2018.

\bibitem[Nachum et~al.(2019)Nachum, Gu, Lee, and Levine]{nachum2019near}
Nachum, O., Gu, S.~S., Lee, H., and Levine, S.
\newblock Near-optimal representation learning for hierarchical reinforcement learning.
\newblock In \emph{International Conference on Learning Representations (ICLR)}, 2019.

\bibitem[Newey \& Powell(1987)Newey and Powell]{newey1987asymmetric}
Newey, W. and Powell, J.~L.
\newblock Asymmetric least squares estimation and testing.
\newblock \emph{Econometrica}, 55\penalty0 (4):\penalty0 819--847, 1987.

\bibitem[Park et~al.(2024{\natexlab{a}})Park, Oh, and Kim]{park2024ngte}
Park, J., Oh, S., and Kim, Y.
\newblock Novelty-aware graph traversal and expansion for hierarchical reinforcement learning.
\newblock In \emph{Conference on Information and Knowledge Management (CIKM)}, 2024{\natexlab{a}}.

\bibitem[Park et~al.(2023)Park, Ghosh, Eysenbach, and Levine]{park2023hiql}
Park, S., Ghosh, D., Eysenbach, B., and Levine, S.
\newblock Hiql: Offline goal-conditioned rl with latent states as actions.
\newblock In \emph{Neural Information Processing Systems (NeurIPS)}, 2023.

\bibitem[Park et~al.(2024{\natexlab{b}})Park, Frans, Levine, and Kumar]{park2024value}
Park, S., Frans, K., Levine, S., and Kumar, A.
\newblock Is value learning really the main bottleneck in offline rl?
\newblock In \emph{Neural Information Processing Systems (NeurIPS)}, 2024{\natexlab{b}}.

\bibitem[Park et~al.(2024{\natexlab{c}})Park, Kreiman, and Levine]{park2024foundation}
Park, S., Kreiman, T., and Levine, S.
\newblock Foundation policies with hilbert representations.
\newblock In \emph{International Conference on Machine Learning (ICML)}, 2024{\natexlab{c}}.

\bibitem[Park et~al.(2025{\natexlab{a}})Park, Frans, Eysenbach, and Levine]{park2025ogbench}
Park, S., Frans, K., Eysenbach, B., and Levine, S.
\newblock Ogbench: Benchmarking offline goal-conditioned rl.
\newblock In \emph{International Conference on Learning Representations (ICLR)}, 2025{\natexlab{a}}.

\bibitem[Park et~al.(2025{\natexlab{b}})Park, Li, and Levine]{park2025flow}
Park, S., Li, Q., and Levine, S.
\newblock Flow q-learning.
\newblock In \emph{International Conference on Machine Learning (ICML)}, 2025{\natexlab{b}}.

\bibitem[Shin \& Kim(2023)Shin and Kim]{shin2023guide}
Shin, W. and Kim, Y.
\newblock Guide to control: Offline hierarchical reinforcement learning using subgoal generation for long-horizon and sparse-reward tasks.
\newblock In \emph{International Joint Conference on Artificial Intelligence (IJCAI)}, 2023.

\bibitem[Sikchi et~al.(2024)Sikchi, Chitnis, Touati, Geramifard, Zhang, and Niekum]{sikchi2024smore}
Sikchi, H., Chitnis, R., Touati, A., Geramifard, A., Zhang, A., and Niekum, S.
\newblock Score models for offline goal-conditioned reinforcement learning.
\newblock In \emph{International Conference on Learning Representations (ICLR)}, 2024.

\bibitem[Sobal et~al.(2025)Sobal, Zhang, Cho, Balestriero, Rudner, and LeCun]{sobal2025latent}
Sobal, V., Zhang, W., Cho, K., Balestriero, R., Rudner, T. G.~J., and LeCun, Y.
\newblock Learning from reward-free offline data: A case for planning with latent dynamics models.
\newblock \emph{ArXiv, abs/2502.14819}, 2025.

\bibitem[Sutton et~al.(1999)Sutton, Precup, and Singh]{sutton1999between}
Sutton, R.~S., Precup, D., and Singh, S.
\newblock Between mdps and semi-mdps: A framework for temporal abstraction in reinforcement learning.
\newblock In \emph{International Conference on Machine Learning (ICML)}, 1999.

\bibitem[Todorov et~al.(2012)Todorov, Erez, and Tassa]{todorov2012mujoco}
Todorov, E., Erez, T., and Tassa, Y.
\newblock Mujoco: A physics engine for model-based control.
\newblock In \emph{International Conference on Intelligent Robots and Systems (IROS)}, 2012.

\bibitem[Vezhnevets et~al.(2017)Vezhnevets, Osindero, Schaul, Heess, Jaderberg, Silver, and Kavukcuoglu]{vezhnevets2017feudal}
Vezhnevets, A.~S., Osindero, S., Schaul, T., Heess, N., Jaderberg, M., Silver, D., and Kavukcuoglu, K.
\newblock Feudal networks for hierarchical reinforcement learning.
\newblock In \emph{International Conference on Machine Learning (ICML)}, 2017.

\bibitem[Wang et~al.(2024)Wang, Yang, Chen, Sun, Fang, and Montana]{wang2024goplan}
Wang, M., Yang, R., Chen, X., Sun, H., Fang, M., and Montana, G.
\newblock Goplan: Goal-conditioned offline reinforcement learning by planning with learned models.
\newblock \emph{Transactions on Machine Learning Research (TMLR)}, 2024.

\bibitem[Wang et~al.(2023)Wang, Torralba, Isola, and Zhang]{wang2023optimal}
Wang, T., Torralba, A., Isola, P., and Zhang, A.
\newblock Optimal goal-reaching reinforcement learning via quasimetric learning.
\newblock In \emph{International Conference on Machine Learning (ICML)}, 2023.

\bibitem[Wu et~al.(2019)Wu, Tucker, and Nachum]{wu2019behavior}
Wu, Y., Tucker, G., and Nachum, O.
\newblock Behavior regularized offline reinforcement learning.
\newblock \emph{ArXiv, abs/1911.11361}, 2019.

\bibitem[Yang et~al.(2023)Yang, Lin, Ma, Hu, Zhang, and Zhang]{yang2023essential}
Yang, R., Lin, Y., Ma, X., Hu, H., Zhang, C., and Zhang, T.
\newblock What is essential for unseen goal generalization of offline goal-conditioned rl?
\newblock In \emph{International Conference on Machine Learning (ICML)}, 2023.

\bibitem[Yoon et~al.(2024)Yoon, Lee, Ahn, and Ok]{yoon2024breadth}
Yoon, Y., Lee, G., Ahn, S., and Ok, J.
\newblock Breadth-first exploration on adaptive grid for reinforcement learning.
\newblock In \emph{International Conference on Machine Learning (ICML)}, 2024.

\bibitem[Zakka et~al.(2022)Zakka, Tassa, and {MuJoCo Menagerie Contributors}]{zakka2022mujoco}
Zakka, K., Tassa, Y., and {MuJoCo Menagerie Contributors}.
\newblock {MuJoCo} menagerie: A collection of high-quality simulation models for {MuJoCo}, 2022.
\newblock URL \url{https://github.com/google-deepmind/mujoco_menagerie}.

\bibitem[Zhang et~al.(2020)Zhang, Guo, Tan, Hu, and Chen]{zhang2020generating}
Zhang, T., Guo, S., Tan, T., Hu, X., and Chen, F.
\newblock Generating adjacency-constrained subgoals in hierarchical reinforcement learning.
\newblock In \emph{Neural Information Processing Systems (NeurIPS)}, 2020.

\bibitem[Zheng et~al.(2024)Zheng, Eysenbach, Walke, Yin, Fang, Salakhutdinov, and Levine]{zheng2024stabilizing}
Zheng, C., Eysenbach, B., Walke, H., Yin, P., Fang, K., Salakhutdinov, R., and Levine, S.
\newblock Stabilizing contrastive rl: Techniques for robotic goal reaching from offline data.
\newblock In \emph{International Conference on Learning Representations (ICLR)}, 2024.

\end{thebibliography}
\bibliographystyle{icml2025}

\appendix
\onecolumn
\section{Graph Construction}
\label{app:graph_construction}

\begin{algorithm}[h!t!]
    \caption{TD-Aware Graph Construction}
    \begin{algorithmic}
        \STATE \textbf{Input:} Offline Dataset $\mathcal{D}$, TDR $\psi$, Temporal Distance Threshold $H_{\text{TD}}$, TE Threshold $\theta_{TE}^{\text{thresh}}$
        \STATE \textbf{Output:} Graph $\mathcal{H}_{\text{graph}} = (\mathcal{V}, \mathcal{E})$
        \vspace{15pt}  
        \STATE \textbf{// TE Filtering}
        \STATE Initialize TDR state set: $\mathcal{H} \gets \emptyset$: 
        \FOR{each trajectory $\tau \in \mathcal{D}$}
            \FOR{each state $s_{\text{cur}} \in \tau$}
                \STATE $h_{\text{cur}} = \psi(s_{\text{cur}})$
                \STATE Optimal state: $h_{\text{opt}} = \psi(\mathcal{F}(s_{\text{cur}}, H_{\text{TD}}))$ // Eq.~\eqref{eq:opt_future_state_fn} 
                \STATE Actual reached state: $h_{\text{reached}} = \psi(s_{\text{cur} + H_{\text{TD}}})$
                \STATE $\theta_\text{TE} = \text{cos}(h_{\text{opt}}-h_{\text{cur}}, h_{\text{reached}}-h_{\text{cur}})$
                \IF{$\theta_\text{TE} \geq \theta_{TE}^{\text{thresh}}$}
                    \STATE $\mathcal{H} \gets \mathcal{H} \cup \{h_{\text{cur}}\}$
                \ENDIF
            \ENDFOR
        \ENDFOR
        \vspace{15pt} 
        \STATE \textbf{// TD-Aware Clustering}
        \STATE $\mathcal{V} \gets \{h_1\}$
        \STATE $\mathcal{C}_1 \gets \{h_1\}$
        \FOR{each state $h_i \in \mathcal{H}, i > 1$}
            \STATE Find the nearest center: $h_c = \arg\min_{h \in \mathcal{V}} \|h_i - h\|_2$
            \IF{$\|h_i - h_c\|_2 > H_{\text{TD}} / 2$}  
                \STATE Create a new cluster: $\mathcal{C}_i \gets \{h_i\}$
                \STATE Insert a new cluster center node: $\mathcal{V} \gets \mathcal{V} \cup \{h_i\}$
            \ELSE
                \STATE Assign $h_i$ to existing cluster: $\mathcal{C}_c \gets \mathcal{C}_c \cup \{h_i\}$
            \ENDIF
        \ENDFOR
        \STATE Reset node set: $\mathcal{V} \gets \emptyset$
        \FOR{each existing cluster $\mathcal{C}_c$}
            \STATE Update cluster center: $h_c = \frac{1}{|\mathcal{C}_c|} \sum_{h_i \in \mathcal{C}_c} h_i$
            \STATE Insert updated center node: $\mathcal{V} \gets \mathcal{V} \cup \{h_c\}$
        \ENDFOR
        \vspace{15pt} 
        \STATE \textbf{// Graph Edge Connection}
        \STATE Initialize edge set $\mathcal{E} \gets \emptyset$
        \FOR{each pair of nodes $(v_i, v_j) \in \mathcal{V}$}
            \STATE Compute distance: $d_{ij} = \|v_i - v_j\|_2$
            \IF{$d_{ij} \leq H_{\text{TD}}$}  
                \STATE $\mathcal{E} \gets \mathcal{E} \cup \{(v_i, v_j)\}$
            \ENDIF
        \ENDFOR
\end{algorithmic}
\end{algorithm}

\textbf{Computational complexity.}
The time complexity of TE (Temporal Efficiency) filtering is $\mathcal{O}(N)$, where $N$ is the total number of states in the dataset. After filtering, we perform TD-aware clustering on $M$ selected high-TE states, which incurs a time complexity of $\mathcal{O}(M^2)$. Finally, constructing the graph edges by computing pairwise distances between nodes requires $\mathcal{O}(|\mathcal{V}|^2)$, where $|\mathcal{V}|$ denotes the number of graph nodes.

\section{Additional Results}
\label{app:additional_results}

\vspace{0.5em}

\begin{figure}[H]
\centering
\includegraphics[width=0.98\linewidth]{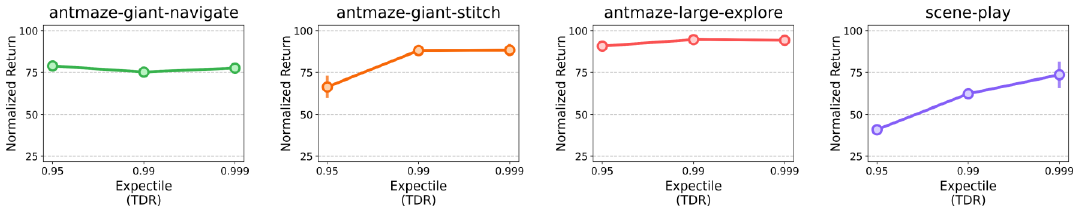}
\caption{\textbf{Ablation on TDR expectile.} Performance across different values of the Temporal Distance Representation (TDR) expectile. The figure shows that sufficiently high expectile values lead to robust performance across most environments.}
\label{fig:tdr_expectile_ablation}
\end{figure}

\begin{figure}[H]
\centering
\includegraphics[width=0.98\linewidth]{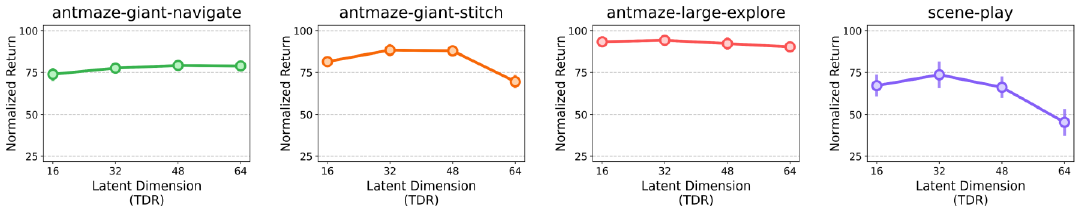}
\caption{\textbf{Ablation on TDR dimension.} Performance across different latent dimensions of the Temporal Distance Representation (TDR). Dimensions between 16 and 48 generally lead to robust performance across tasks.}
\label{fig:tdr_dimension_ablation}
\end{figure}

\begin{figure}[H]
\centering
\includegraphics[width=0.98\linewidth]{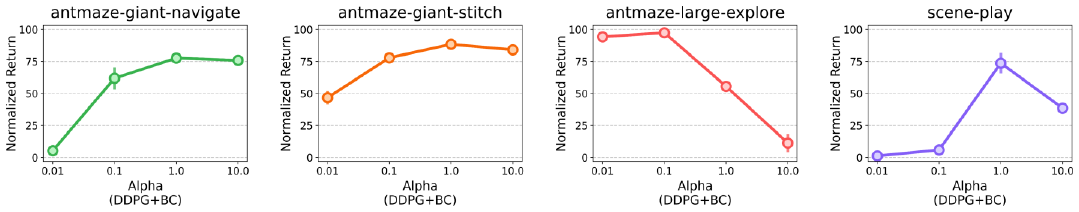}
\caption{\textbf{Ablation on BC coefficient.} Performance across different $\alpha$ values in deep deterministic policy gradient and behavioral cloning (DDPG+BC). The optimal $\alpha$ depends on dataset quality, and selecting it within an appropriate range can improve task performance.}
\label{fig:bc_coefficient_ablation}
\end{figure}

\textbf{Ablation on TDR expectile.}
Figure~\ref{fig:tdr_expectile_ablation} shows how varying the value of the Temporal Distance Representation (TDR) expectile affects performance across tasks.
The expectile coefficient plays a key role in capturing temporal distance (i.e., the minimum number of steps required to transition from one state to another in the original state space). We found that using a sufficiently high expectile value (e.g., 0.95 or above) led to robust performance across most environments.

\vspace{0.5em}

\textbf{Ablation on TDR dimension.}
Figure~\ref{fig:tdr_dimension_ablation} shows how varying the latent dimension of the TDR affects performance across tasks. We observe that dimensions between 16 and 48 generally lead to robust performance. Based on this analysis, we fix the latent dimension to 32 for all experiments to balance representational capacity and stability.

\vspace{0.5em}

\textbf{Ablation on BC coefficient.}
Figure~\ref{fig:bc_coefficient_ablation} shows how varying the coefficient $\alpha$ in the deep deterministic policy gradient and behavioral cloning (DDPG+BC) objective affects performance across tasks.
The $\alpha$ coefficient controls the strength of the BC regularizer. As shown in the figure, the effect of $\alpha$ on performance varies depending on the dataset quality. For instance, in the antmaze-large-explore environment, which contains many random movements and noisy actions, using a smaller $\alpha$ (i.e., assigning less weight to the BC term) leads to better performance.

\section{Environments and Datasets}
\label{app:environments_and_datasets}

\vspace{0.5em}

\begin{figure}[H]
    \centering
    \includegraphics[width=0.98\linewidth, trim=0 0 0 50, clip]{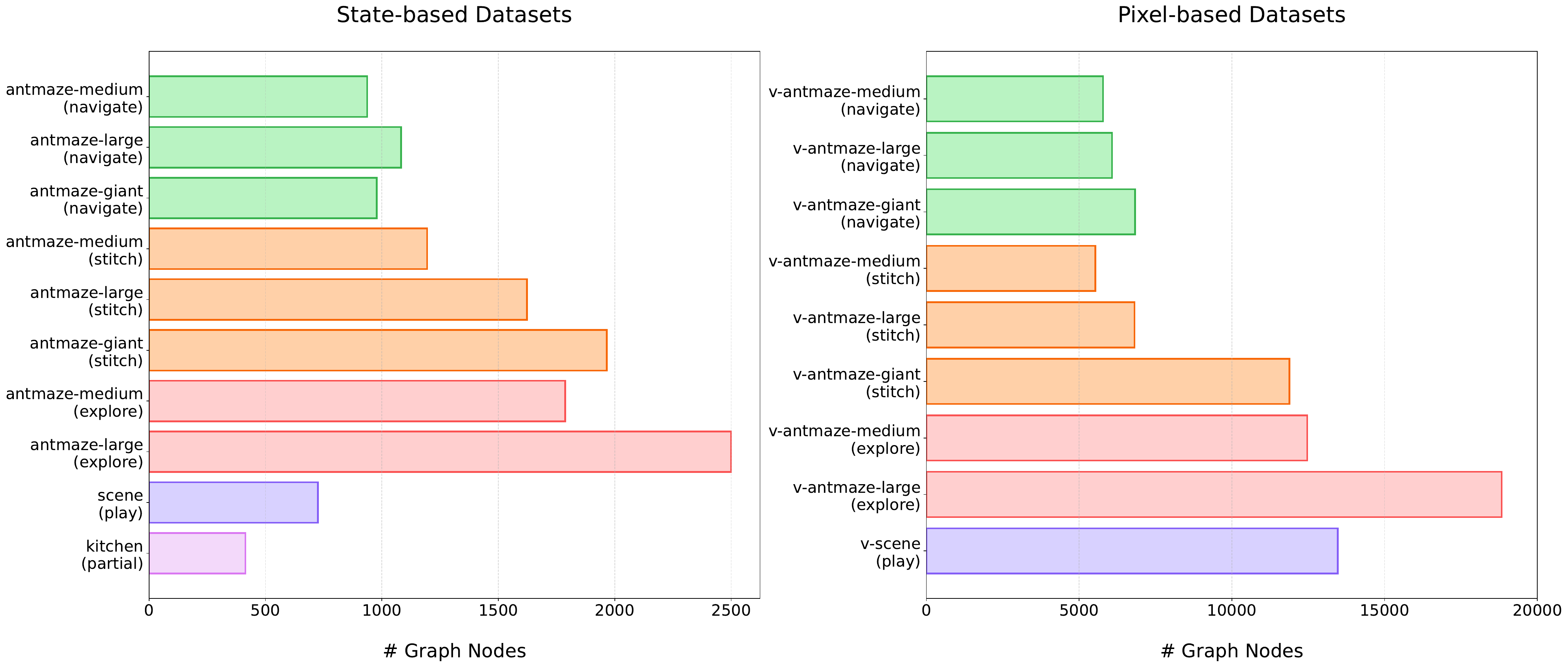}
    \caption{\textbf{Number of graph nodes per dataset.} 
    \textit{(left)}: state-based environments. \textit{(right)}: pixel-based environments.}
    \label{fig:graph_node_statistics}
\end{figure}

\begin{table}[H]
    \centering
    \caption{Summary of offline datasets and environments.}
    \label{tab:dataset_statistics}
    \resizebox{0.95\linewidth}{!}{
    \begin{tabular}{@{}llccccc@{}}
        \toprule
        \textbf{Dataset} & \textbf{Dataset Type} & \textbf{\# Transitions} & \textbf{Maximum Episode Length} & \textbf{State Dim} & \textbf{Action Dim} \\ 
        \midrule
        antmaze-medium-navigate          & Locomotion   & 1M & 1000 & 29 & 8 \\
        antmaze-large-navigate           & Locomotion   & 1M & 1000 & 29 & 8 \\
        antmaze-giant-navigate           & Locomotion   & 1M & 2000 & 29 & 8 \\
        antmaze-medium-stitch            & Stitching    & 1M &  200 & 29 & 8 \\
        antmaze-large-stitch             & Stitching    & 1M &  200 & 29 & 8 \\
        antmaze-giant-stitch             & Stitching    & 1M &  200 & 29 & 8 \\
        antmaze-medium-explore           & Exploratory  & 5M &  500 & 29 & 8 \\
        antmaze-large-explore            & Exploratory  & 5M &  500 & 29 & 8 \\
        scene-play                       & Manipulation & 1M & 1000 & 40 & 5 \\
        kitchen-partial                  & Manipulation & 0.13M &  526 & 30 & 9 \\
        \midrule
        visual-antmaze-medium-navigate   & Locomotion   & 1M & 1000 & $64\!\times\!64\!\times\!3$ & 8 \\
        visual-antmaze-large-navigate    & Locomotion   & 1M & 1000 & $64\!\times\!64\!\times\!3$ & 8 \\
        visual-antmaze-giant-navigate    & Locomotion   & 1M & 2000 & $64\!\times\!64\!\times\!3$ & 8 \\
        visual-antmaze-medium-stitch     & Stitching    & 1M &  200 & $64\!\times\!64\!\times\!3$ & 8 \\
        visual-antmaze-large-stitch      & Stitching    & 1M &  200 & $64\!\times\!64\!\times\!3$ & 8 \\
        visual-antmaze-giant-stitch      & Stitching    & 1M &  200 & $64\!\times\!64\!\times\!3$ & 8 \\
        visual-antmaze-medium-explore    & Exploratory  & 5M &  500 & $64\!\times\!64\!\times\!3$ & 8 \\
        visual-antmaze-large-explore     & Exploratory  & 5M &  500 & $64\!\times\!64\!\times\!3$ & 8 \\
        visual-scene-play                & Manipulation & 1M & 1000 & $64\!\times\!64\!\times\!3$ & 5 \\
        \bottomrule
    \end{tabular}
    }
\end{table}

\vspace{0.5em}

\textbf{Graph node statistics.} We report the number of graph nodes constructed for each environment, as shown in Figure~\ref{fig:graph_node_statistics}. GAS constructs a compact graph by performing TD-aware clustering on high-TE states selected via Temporal Efficiency (TE) filtering. This substantial reduction in the number of nodes significantly improves overall efficiency in task planning and execution (Algorithm~\ref{algo:task_planning_execution}). For example, in the antmaze-large-explore, only 2,499 nodes are selected from 5 million transitions, corresponding to approximately 0.05\% of all states in the dataset. In the visual-antmaze-large-explore, 18,830 nodes are selected, representing approximately 0.38\% of all states.

\vspace{0.5em}

\textbf{Dataset statistics.}
Table~\ref{tab:dataset_statistics} summarizes the key statistics of offline datasets and environments used in our experiments from OGBench~\cite{park2025ogbench} and D4RL~\cite{fu2020d4rl}. 

\vspace{0.5em}

\noindent\textbf{Environments}
\begin{itemize}[leftmargin=1.0em, itemsep=0.5em, topsep=1.5pt]
    \item \textbf{Antmaze / Visual-Antmaze}: The agent controls a quadruped Ant robot~\cite{brockman2016gym} with 8 degrees of freedom (DoF) to reach a designated goal location in a maze. This environment requires both high-level maze navigation and low-level locomotion skills. It includes three maze sizes: medium, large, and giant, where larger mazes require more long-horizon reasoning~\cite{park2025ogbench}. The state-based observation space consists of 29 dimensions, including the agent's current 2D coordinates (i.e., \textit{x-y} position) and joint-related features. In the pixel-based environments, the agent receives only $64{\times}64{\times}3$ RGB images rendered from a third-person camera viewpoint, and the maze floor is color-coded to enable the agent to infer its location solely from visual inputs. All environments are simulated in the MuJoCo physics engine~\cite{todorov2012mujoco}.

    \item \textbf{Scene / Visual-Scene}: The agent manipulates everyday objects (e.g., cube blocks, a drawer, a window, and two button locks) using a 6-DoF UR5e robot arm~\cite{zakka2022mujoco} to achieve specific configurations. This environment requires sequential reasoning and diverse manipulation skill composition~\cite{park2025ogbench}. For example, to place a cube in the drawer, the agent must first press a button to unlock it, then open the drawer before inserting the cube. This process requires the agent to make precise sequential decisions across diverse manipulation skills. The state-based observation space consists of 40 dimensions, including joint and object-related features. In the pixel-based environments, the agent receives $64{\times}64{\times}3$ RGB images, with the robot arm rendered semi-transparent and colors adjusted to ensure full observability, without requiring high-resolution inputs or multiple camera views. All environments are based on the MuJoCo physics engine~\cite{todorov2012mujoco}.

    \item \textbf{Kitchen}: The agent controls a 9-DoF Franka robotic arm~\cite{gupta2020relay} to achieve four target subtasks, such as opening the microwave, moving the kettle, turning on the light switch, and sliding the cabinet door~\cite{fu2020d4rl}. Only a small portion of the dataset contains successful trajectories that complete all four subtasks sequentially. The state-based observation space is 30-dimensional, including  joint and object-related features. 
\end{itemize}

\vspace{0.5em}

\noindent\textbf{Dataset Types}
\begin{itemize}[leftmargin=1.0em, itemsep=0.5em, topsep=1.5pt]
  \item \textbf{Locomotion:} Evaluates long-horizon reasoning ability. The agent must navigate toward a goal state over multiple steps, requiring long-term planning. The dataset~\cite{park2025ogbench} contains long-horizon goal-reaching trajectories collected by a noisy expert policy~\cite{haarnoja2018sac} navigating to randomly sampled goals.
  \item \textbf{Stitching:} Evaluates the agent’s ability to stitch together partial trajectories by combining previously unconnected segments into an optimal plan. The dataset~\cite{park2025ogbench} contains short goal-reaching trajectories collected by a noisy expert policy~\cite{haarnoja2018sac} navigating to randomly sampled goals.
  \item \textbf{Exploratory:} Evaluates the ability to learn from suboptimal datasets. The agent must utilize low-quality, high-coverage data to learn an effective policy. The dataset~\cite{park2025ogbench} is collected by commanding the noisy expert policy~\cite{haarnoja2018sac} with random directions re-sampled every 10 steps under high action noise.
  \item \textbf{Manipulation:} Evaluates sequential decision-making ability that requires reasoning about prior actions when planning future actions. The play-style dataset~\cite{lynch2019learning} is collected using open-loop, non-Markovian scripted policies that randomly interact with  diverse everyday objects~\cite{park2025ogbench}. The partial dataset~\cite{fu2020d4rl} is derived from a set of unstructured and unsegmented human demonstrations, originally collected using the PUPPET MuJoCo VR system~\cite{kumar2015mujoco, gupta2020relay}.
\end{itemize}

\vspace{0.5em}

\textbf{Goal Specification for Evaluation}
We follow the goal specification protocol of OGBench~\cite{park2025ogbench}, where each task provides five predefined state-goal pairs. For the kitchen-partial environment~\cite{fu2020d4rl, gupta2020relay}, we construct a single fixed goal state. Specifically, we extract the joint configuration from a dataset observation and combine it with the object configuration corresponding to the target goal state, as provided by the environment.  

\vspace{0.5em}

\textbf{Evaluation Metric}
We use normalized return as the primary evaluation metric, where the maximum achievable reward is normalized to 100. The reward structure varies across environments. Antmaze provides a sparse reward of +1 only when the agent reaches the goal state. Scene assigns a reward of +1 only if all required subtasks are completed. Kitchen provides +1 for each successfully completed subtask, with a maximum cumulative reward of 4. All scores are linearly scaled to the range [0, 100]~\cite{fu2020d4rl, park2025ogbench}.

\section{Implementation Details}
\label{app:implementation_details}

\vspace{0.5em}

\textbf{Implementation}
Our implementations of GAS and seven baselines are based on JAX~\cite{bradbury2018jax}. We run our experiments on an internal cluster consisting of RTX 3090 GPUs. We also test and tune different design choices and hyperparameters for both GAS and baselines, largely following prior work~\cite{park2023hiql, park2024foundation, park2025ogbench}. To ensure fair comparison, we use a unified set of model architectures and hyperparameter values across GAS and baselines (e.g., learning rate, discount factors, and batch sizes) whenever possible. Our implementations are publicly available at the following repository: \url{https://github.com/qortmdgh4141/GAS}.

\vspace{0.5em}

\textbf{Baselines} We compare GAS against of four offline goal-conditioned algorithms (GCBC, GCIQL, QRL, CRL) and three hierarchical algorithms (HGCBC, HIQL, HHILP). In goal-conditioned offline RL, we include GCBC, a simple goal-conditioned behavioral cloning method, and GCIQL~\cite{kostrikov2022offline, park2023hiql}, which approximates the optimal value function using expectile regression. We also evaluate QRL~\cite{wang2023optimal}, which learns a quasimetric value function with a dual objective, and CRL~\cite{eysenbach2022contrastive}, which applies contrastive learning to estimate Monte Carlo value functions and perform one-step policy improvement. For hierarchical approaches, we consider HGCBC~\cite{ghosh2021learning, park2023hiql}, an extension of GCBC~\cite{gupta2020relay, ghosh2021learning} incorporating two-level policies, and HHILP, a hierarchical extension of HILP~\cite{park2023hiql, park2024foundation} that utilizes Temporal Distance Representation (TDR). Finally, we benchmark HIQL~\cite{park2023hiql}, which extends GCIQL to a hierarchical structure and generates subgoals in latent space.

\vspace{0.5em}

\textbf{Temporal Distance Representation (TDR).}
We follow the goal relabeling strategy proposed in~\cite{andrychowicz2017hindsight, park2023hiql, park2024foundation}, with the exception that we exclude the case $s = g$, as our parameterization guarantees $V(s, s) = 0$. Specifically, we sample $g$ either from a geometric distribution over future states within the same trajectory (with probability 0.625), or uniformly from the entire dataset (with probability 0.375). These values are inherited from the original hyperparameter settings~\cite{park2024foundation}, which ensure that $s = g$ is never sampled by assigning the probability mass to both of the two sampling schemes mentioned above.

\vspace{0.5em}

\textbf{Hyperparameters} We provide a common list of hyperparameters in Table~\ref{table:common_hyperparameter} and task-specific hyperparameters in Table~\ref{table:task_specific_hyperparameter}. We apply layer normalization~\cite{layernorm} to all MLP layers. For pixel-based environments, we adopt the Impala CNN~\cite{espeholt2018impala} to process image inputs. While most components use 512-dimensional output features, we reduce the output dimension to 32 for the Temporal Distance Representation (TDR) to balance representational capacity and stability, as discussed in Appendix~\ref{app:additional_results}. Following prior work~\cite{park2023hiql, park2024foundation, park2025ogbench}, we do not share encoders across components. As a result, in pixel-based environments, we use four separate CNN encoders for TDR, the Q-function, the value function, and the low-level policy. We also apply random crop augmentation~\cite{kostrikov2021image} with a probability of 0.5 to mitigate overfitting~\cite{zheng2024stabilizing}.

\vspace{0.5em}

\begin{table}[H] 
    \centering   
    \caption{Common hyperparameters used across all datasets.}
    \label{table:common_hyperparameter}
    \resizebox{0.77\linewidth}{!}{
    \begin{tabular}{lc}
        \toprule
        \textbf{Hyperparameter} & \textbf{Value} \\
        \midrule
        Image representation architecture (pixel-based) & Impala CNN~\cite{espeholt2018impala} \\
        Image augmentation probability (pixel-based) & 0.5~\cite{kostrikov2021image} \\
        TDR MLP dimension & (512, 512, 512) \\
        Value MLP dimensions & (512, 512, 512) \\
        Policy MLP dimensions & (512, 512, 512) \\
        TDR Latent dimension & 32 \\
        TDR normalization & LayerNorm~\cite{layernorm} \\
        Value normalization & LayerNorm \\
        Nonlinearity & GELU~\cite{gelu} \\
        Value expectile & 0.7 \\
        Target network smoothing coefficient & 0.005 \\
        Learning rate & 0.0003 \\
        Optimizer & Adam~\cite{kingma2015adam} \\
        \bottomrule
    \end{tabular}
    }
\end{table}

\begin{table}[H]
    \centering
    \caption{Task-specific hyperparameters for each dataset.}
    \label{table:task_specific_hyperparameter}
    \resizebox{0.99\linewidth}{!}{
    \begin{tabular}{lccccccc}
        \toprule
        \textbf{Dataset} & \textbf{\# Gradient Steps} & \textbf{Batch Size} & \textbf{Discount Factor} & \textbf{TDR Expectile} & $\theta_{TE}^{\text{thresh}}$ & \textbf{$H_\text{TD}$} & $\boldsymbol{\alpha}$ \\
        \midrule
        antmaze-medium-navigate & 1000000 & 1024 & 0.99  & 0.999 & 0.99 & 8  & 1 \\
        antmaze-large-navigate  & 1000000 & 1024 & 0.99  & 0.999 & 0.99 & 8  & 1 \\
        antmaze-giant-navigate  & 1000000 & 1024 & 0.995 & 0.999 & 0.99 & 8  & 1 \\
        antmaze-medium-stitch   & 1000000 & 1024 & 0.99  & 0.999 & 0.99 & 8  & 1 \\
        antmaze-large-stitch    & 1000000 & 1024 & 0.99  & 0.999 & 0.99 & 8  & 1 \\
        antmaze-giant-stitch    & 1000000 & 1024 & 0.995 & 0.999 & 0.99 & 8  & 1 \\
        antmaze-medium-explore  & 1000000 & 1024 & 0.99  & 0.999 & 0.99 & 8  & 0.01 \\
        antmaze-large-explore   & 1000000 & 1024 & 0.99  & 0.999 & 0.99 & 8  & 0.01 \\
        scene-play              & 1000000 & 1024 & 0.99  & 0.999 & 0.99 & 48 & 1 \\
        kitchen-partial         & 500000  & 1024  & 0.99  & 0.95  & 0.9  & 48 & 10 \\
        \midrule        
        visual-antmaze-medium-navigate & 500000 & 256 & 0.99 & 0.95 & 0.9 & 8  & 1 \\
        visual-antmaze-large-navigate  & 500000 & 256 & 0.99 & 0.95 & 0.9 & 8  & 1 \\
        visual-antmaze-giant-navigate  & 500000 & 256 & 0.995 & 0.95 & 0.9 & 8  & 1 \\
        visual-antmaze-medium-stitch   & 500000 & 256 & 0.99 & 0.95 & 0.9 & 8  & 1 \\
        visual-antmaze-large-stitch    & 500000 & 256 & 0.99 & 0.95 & 0.9 & 8  & 1 \\
        visual-antmaze-giant-stitch    & 500000 & 256 & 0.995 & 0.95 & 0.9 & 8  & 1 \\
        visual-antmaze-medium-explore  & 500000 & 256 & 0.99 & 0.95 & 0.9 & 8  & 0.01 \\
        visual-antmaze-large-explore   & 500000 & 256 & 0.99 & 0.95 & 0.9 & 8  & 0.01 \\
        visual-scene-play              & 500000 & 256 & 0.99 & 0.95 & 0.9 & 24 & 1 \\
        \bottomrule
    \end{tabular}
    }
\end{table}

\vspace{0.5em}

\section{Subgoal Visualizations}

\vspace{0.5em}

\begin{figure}[H]
\centering
\includegraphics[width=0.99\linewidth]{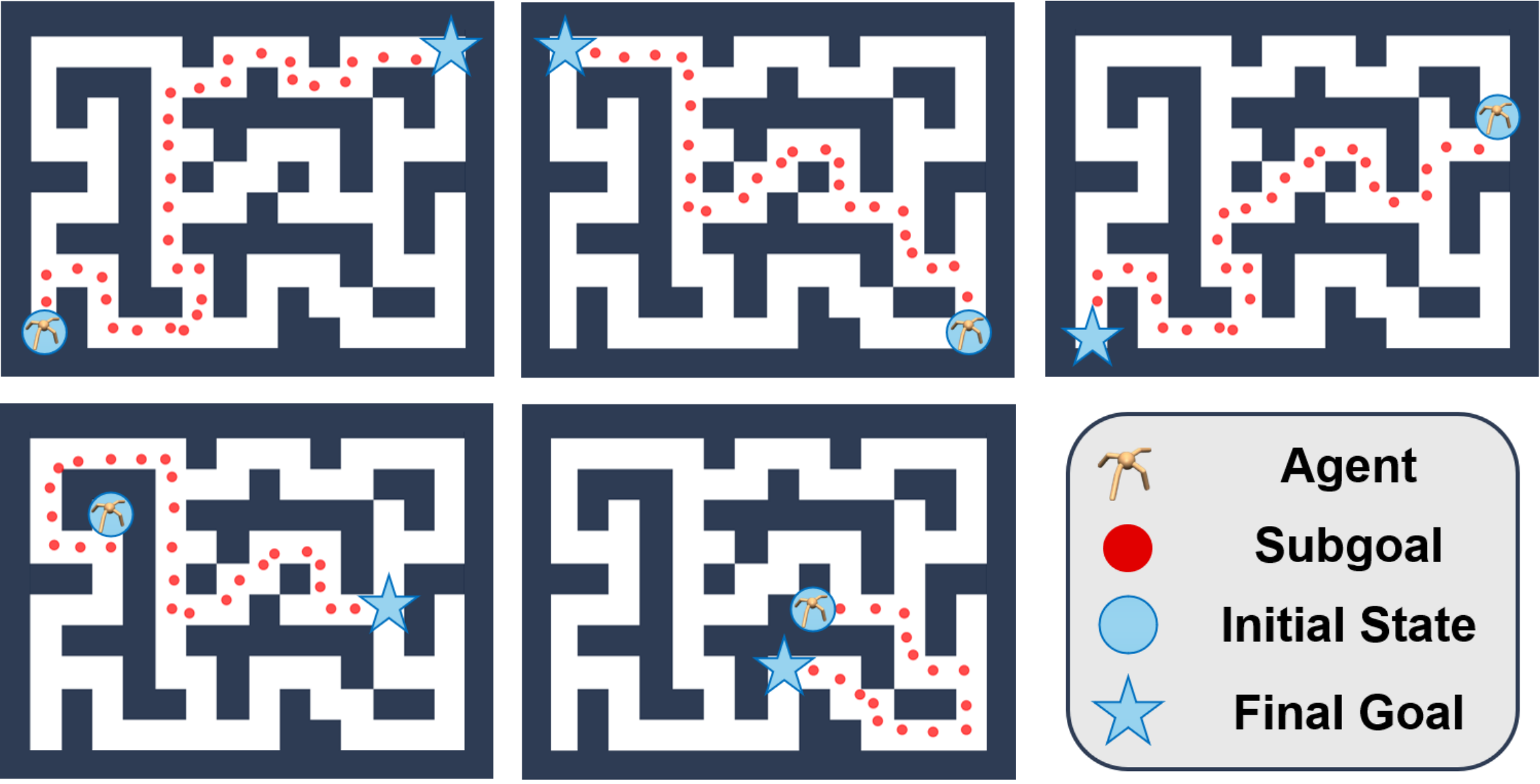}
\caption{\textbf{Shortest path computed on the TD-aware graph from antmaze-giant-stitch.} The graph is constructed in a latent space, but for visualization purposes, each node embedding is projected onto a 2D plane using approximate 2D coordinates (i.e., \textit{x-y} position).}
\end{figure}
\end{document}